\documentclass[runningheads]{llncs}

\usepackage{eccv}

\usepackage{eccvabbrv}

\usepackage{graphicx}
\usepackage{booktabs}

\usepackage[accsupp]{axessibility}  % Improves PDF readability for those with disabilities.

\usepackage{hyperref}
\hypersetup{hidelinks}

\usepackage{orcidlink}
\usepackage{graphicx}
\usepackage{caption}
\usepackage{subcaption}

\usepackage{algorithm}
\usepackage{algpseudocode}
\usepackage{todonotes}
\usepackage{wrapfig}
\usepackage{tabularx, makecell}
\newcommand\OurMethod{PLSR}
\usepackage{filecontents}
\usepackage{multibib}
\newcites{appendix}{Reference}

\usepackage{xcolor}
\usepackage{soul}
\usepackage{caption}

\sethlcolor{cyan}

\newif\ifmarkchanges
\markchangesfalse

\DeclareRobustCommand{\markhl}[1]{%
  \ifmarkchanges
    \hl{#1}%
  \else
    #1%
  \fi
}
\definecolor{ao}{rgb}{0.0, 0.0, 1.0}
\definecolor{airforceblue}{rgb}{0.36, 0.54, 0.66}
\definecolor{ceruleanblue}{rgb}{0.16, 0.32, 0.75}
\definecolor{cerulean}{rgb}{0.0, 0.48, 0.65}
\definecolor{celestialblue}{rgb}{0.29, 0.59, 0.82}
\definecolor{azure(colorwheel)}{rgb}{0.0, 0.5, 1.0}
\definecolor{babyblue}{rgb}{0.54, 0.81, 0.94}
\definecolor{babyblueeyes}{rgb}{0.63, 0.79, 0.95}
\definecolor{ballblue}{rgb}{0.13, 0.67, 0.8}

\definecolor{asparagus}{rgb}{0.53, 0.66, 0.42}
\definecolor{ao(english)}{rgb}{0.0, 0.5, 0.0}
\definecolor{applegreen}{rgb}{0.55, 0.71, 0.0}
\definecolor{armygreen}{rgb}{0.29, 0.33, 0.13}
\definecolor{gray-asparagus}{rgb}{0.27, 0.35, 0.27}
\definecolor{green(ryb)}{rgb}{0.4, 0.69, 0.2}

\definecolor{amethyst}{rgb}{0.6, 0.4, 0.8}
\definecolor{antiquefuchsia}{rgb}{0.57, 0.36, 0.51}
\definecolor{blue-violet}{rgb}{0.54, 0.17, 0.89}
\definecolor{brightlavender}{rgb}{0.75, 0.58, 0.89}
\definecolor{brightube}{rgb}{0.82, 0.62, 0.91}
\definecolor{brilliantlavender}{rgb}{0.96, 0.73, 1.0}

\definecolor{amber}{rgb}{1.0, 0.75, 0.0}
\definecolor{amber(sae/ece)}{rgb}{1.0, 0.49, 0.0}
\definecolor{atomictangerine}{rgb}{1.0, 0.6, 0.4}
\definecolor{burntorange}{rgb}{0.8, 0.33, 0.0}
\definecolor{burntsienna}{rgb}{0.91, 0.45, 0.32}
\definecolor{cadmiumorange}{rgb}{0.93, 0.53, 0.18}
\definecolor{carrotorange}{rgb}{0.93, 0.57, 0.13}
\definecolor{chocolate(web)}{rgb}{0.82, 0.41, 0.12}
\definecolor{chromeyellow}{rgb}{1.0, 0.65, 0.0}
\definecolor{darkorange}{rgb}{1.0, 0.55, 0.0}
\definecolor{darktangerine}{rgb}{1.0, 0.66, 0.07}
\definecolor{deepcarrotorange}{rgb}{0.91, 0.41, 0.17}
\definecolor{deepsaffron}{rgb}{1.0, 0.6, 0.2}
\definecolor{fulvous}{rgb}{0.86, 0.52, 0.0}

\sethlcolor{babyblue}

\usepackage{marvosym}
\usepackage{pdfpages}
\begin{document}

\title{PLSR: Progressive and Localized Super-Resolution of 3D Objects via Localized Latent Voxel Diffusion} 

\titlerunning{PLSR}

% Second Author\inst{2,3}\orcidlink{1111-2222-3333-4444} \and
% Third Author\inst{3}\orcidlink{2222--3333-4444-5555}}
\author{Yuxin Liu \inst{1}\orcidlink{0009-0001-5658-7360} \and Minshan Xie \inst{2}\orcidlink{0000-0002-6288-1611} \and Jiawen Liang \inst{3}\orcidlink{0009-0009-3042-8239} \and Runsong Zhu \inst{1}\thanks{Corresponding author}\Letter\orcidlink{0009-0007-0220-3273} \and Chi-Wing Fu \inst{1}\orcidlink{0000-0002-5238-593X} \and Tien-Tsin Wong \inst{4}\orcidlink{0000-0002-7792-9307}}
% \email{yxliu22@cse.cuhk.edu.hk}
% \affiliation{
%  \institution{The Chinese University of Hong Kong}
%   \country{Hong Kong}
% }

% \author{Minshan Xie}
% \email{msxie92@gmail.com}
% \affiliation{
%  \institution{Guangdong University of Technology}
%   \country{China}
% }

% \author{Chiwing Fu}
% \affiliation{%
%   \institution{The Chinese University of Hong Kong}
%   \country{Hong Kong}
% }
% \email{cwfu@cse.cuhk.edu.hk}

% \author{Tien-Tsin Wong}
% \affiliation{%
%   \institution{Monash University}
%   \city{Melbourne}
%   \country{Australia}}
% \email{tt.wong@monash.edu}

\authorrunning{Liu et al.}

\institute{
The Chinese University of Hong Kong, Hong Kong SAR, China 
\and Guangdong University of Technology, China
\and City University of Hong Kong, Hong Kong SAR, China
\and Monash University, Australia \\
\email{yxliu22@cse.cuhk.edu.hk, zhurunsong@gmail.com}
}

\maketitle

\begin{figure*}[ht]
    \centering
    % First subfigure
    \includegraphics[width=0.9\linewidth]{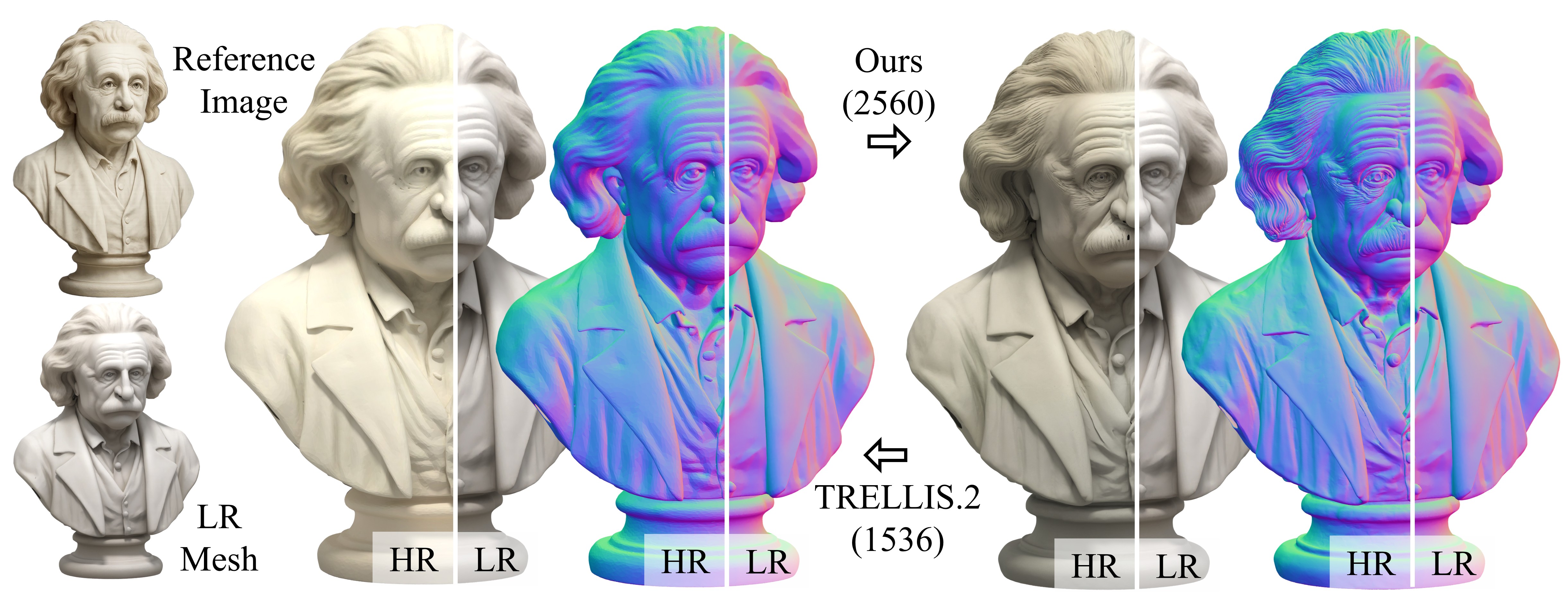} 
    % \missingfigure[figwidth=\linewidth, figheight=6cm]{teaser goes here.}
    \label{fig:teaser}
    
    \caption{We propose \OurMethod, a method that tackles the high-resolution mesh generation problem through \textbf{P}rogressive and \textbf{L}ocalized \textbf{S}uper-\textbf{R}esolution. Our method provides extra resolution increase over state-of-the-art 3D generation methods (e.g., TRELLIS.2). Zoom in to see the fine details.}
\end{figure*}

\begin{abstract}
High-resolution 3D asset generation is vital in various 3D applications.
Existing state-of-the-art diffusion-based models remain constrained by \markhl{fixed resolutions, limiting their ability to produce details.}
In this paper, we tackle the challenge of generating more detailed, higher-resolution 3D objects by introducing a 3D super-resolution (SR) framework built on existing 3D generative foundation models.
% \textbf{}
To this end, we design \markhl{\textbf{PLSR}}, a progressive and localized super-resolution solution to achieve this goal effectively and memory efficiently.
% 
% Technique-wise, we designed an associative input decomposition scheme to break down global super-resolution problems into localized super-resolution problems, and use 
Technically, given a coarse geometry from a pretrained 3D generator, we decompose the global SR task into localized sub-tasks via an \textbf{associative input decomposition} scheme, adapt a flow-based 3D generator into a \textbf{localized super-resolution model} through low-cost finetuning, and unify them in an iterative \textbf{patch-wise denoising pipeline} for seamless high-resolution output.
Experiments on challenging \markhl{objects} show that our approach is able to generate 3D details with new strong fine-detail fidelity while significantly reducing the  
computational cost, offering a new and practical solution for high-resolution 3D asset generation.
%High-resolution 3D asset generation is vital in various 3D applications, yet diffusion-based models remain constrained by cubic memory growth and fixed resolutions. While current methods achieve fidelity on simple objects, they struggle with complex or large-scale assets, and scaling model size or data quickly becomes impractical.
% \phil{
% Existing methods for 3D asset generation often struggle to produce details. 
% Constrained by cubic memory growth and resolution limits, scaling up the model size or data quickly becomes impractical.
%
% This paper introduces a novel} scalable patch-wise

% 

%divide-and-conquer \phil{patch-wise? the word patch seems to match your method better? The divide-and-conquer has a meaning of recursive but it seems that your method doesn't have} 
% framework that decomposes object generation into \phil{associative 2D-3D patches}, each refined by a lightweight super-resolution model. 
% \phil{Especially, by} leveraging \phil{pre-trained} object-level priors, 
%from pre-trained 3D flow models, 
% our method \phil{can} enhance fine-scale structures on textured meshes with minimal modification.
% \phil{this sentence is not specific enough to highlight the technical novelty in the method}
%
% Starting from a base-resolution object, the model applies progressive \phil{remove progressive, since you didn't have it, right?} patchwise \phil{patch-wise} super-resolution conditioned on voxel features and cropped reference images. Predictions are stitched during denoising to ensure global coherence.
%

%
\keywords{3D asset generation \and super-resolution \and progressive \and localized \and diffusion-based 3D models}
\end{abstract}

\section{Introduction}
\label{sec:intro}

High‑resolution 3D geometric models are foundational to numerous industry applications, including computer games, TV/movie production, computer-aided design and fast prototyping. 
Traditionally, creating these models has been time-consuming and requires substantial expertise, and hence limited to trained professionals.
Recent advances in generative artificial intelligence have lowered this barrier by enabling the automatic generation of 3D assets from images.

Diffusion and flow-based models have demonstrated superior performance in generating high-quality 3D content, emerging as a powerful and general solution for 3D asset generation.
In particular, recent methods such as XCube~\cite{ren2024xcube}, {Craftsman}, Hunyuan3D~\cite{zhao2025hunyuan3d}, and TRELLIS.2~\cite{xiang2025native} introduce various techniques to enhance the scalability. 
While significant improvement has been achieved, these models are still typically trained at a fixed resolution in voxel grids or latent spaces, due to the cubic growth in memory and compute costs. On the other hand, users' desired level of details for 3D assets can  vary arbitrarily.

In this paper, we tackle the challenge of generating more detailed, higher-resolution 3D objects by introducing a 3D super-resolution (SR) framework built on existing 3D generative foundation models.
Specifically, a pretrained 3D generator is first employed to produce a coarse global mesh, which is then refined by a dedicated SR model designed to synthesize fine‑grained details. 
% and then a dedicated super-resolution model is designed to synthesize fine-grained details.
%
%Benefiting from the inherent decoupling nature of the SR setting,
Benefiting from this SR setting, our solution naturally inherits the structural priors from the object-level shape to generate challenging fine details, making the learning process more effective.

To do so, we propose a \textbf{progressive} and \textbf{localized} SR \markhl{(PLSR)} framework to generate details with two goals in mind.
First, the super-resolution must primarily rely on local information rather than global information, so that we can adopt a divide-and-conquer strategy to decompose the global SR task into multiple independently processed patches, yielding better memory efficiency for both training and inference.
Second, we want this localized training setting to learn an image-conditioned mapping between LR and HR patches, so that it can be independent of the absolute object resolution. 
\markhl{This allows the SR model to be compatible with objects at any resolution, which is essential for progressive super-resolution.}

Specifically, given a coarse global geometry produced by a pretrained 3D generator, we use an \textbf{associative input  decomposition} scheme to decompose global SR task into localized sub-tasks with aligned local conditions, and adapt an object-level flow-based 3D generator into a \textbf{localized super-resolution model} through low-cost finetuning, which is responsible for processing localized SR tasks. Finally, we design an iterative \textbf{patch-wise denoising pipeline} where global latents are decomposed, processed and stitched per denoising iteration, achieving seamless high-resolution SR using localized processing.

Experiments across various challenging test cases demonstrate that our method consistently \markhl{improves fine-detail fidelity under a practical single-GPU setting.}
% outperforms prior approaches in generative detailization. 
By leveraging localized conditioning and progressive refinement, our framework \markhl{super-resolves coarse 3D objects in the structured} latent~\cite{xiang2025native} domain, achieving higher resolutions with only low-cost fine-tuning and offering a practical solution for scalable 3D super‑resolution.  
Our contributions are summarized as follows:

\begin{itemize}
    \item \textbf{Resource-efficient Super-Resolution (SR) Framework}: We introduce a novel SR framework that achieves substantial resolution enhancement without requiring multi-resolution HR training data or multi-GPU compute, making high-quality 3D generation more accessible.
    \item \textbf{Patch-wise refinement model}: 
    We propose a model design that enables progressive and localized super-resolution without assuming the whole-object generation as in existing methods. 

    \item \textbf{Fine-detail generation}: 
    Experiments on challenging cases demonstrate the superior performance of PLSR to generate fine details.

\end{itemize}
 \section{Related Work}
\label{sec:related_work}
\vspace{-1mm}
\subsection{3D Content Generation}
3D content generation is a long-standing challenge in computer graphics and vision~\cite{zhang2022marching,han2024super,huang2025part}. Early progress was limited by the scale and diversity of classic repositories such as ShapeNet~\cite{chang2015shapenet}. To address data scarcity, zero-shot pipelines~\cite{jain2022zero,michel2022text2mesh} optimized text-prompted 3D representations using 2D language–image priors like CLIP~\cite{radford2021learning}. With the advances of 2D diffusion models~\cite{rombach2022high}, score distillation sampling (SDS) became the dominant paradigm for text-to-3D generation~\cite{poole2023dreamfusion}, inspiring variants such as ProlificDreamer~\cite{wang2023prolificdreamer} and DreamGaussian~\cite{tang2024dreamgaussian}. But SDS-based pipelines remain iterative and suffer from multi-view inconsistency.

The release of large-scale datasets such as Objaverse~\cite{deitke2023objaverse} and Objaverse-XL~\cite{deitke2023objaversexl} enabled feed-forward 3D generation. Leveraging multi-view supervision, LRM~\cite{hong2024lrm} introduced an end-to-end regression framework that predicts triplane NeRF representations~\cite{chan2022efficient} from a single image, while Instant3D~\cite{li2024instant3d} adopts a two-stage pipeline that first synthesizes a “view-compiled” image grid from text prompts and then reconstructs 3D geometry. Recent works improve efficiency and fidelity through Gaussian splatting (LGM~\cite{tang2024lgm}), convolutional backbones (CRM~\cite{wang2024crm}), and enhanced geometric priors (InstantMesh~\cite{xu2024instantmesh}).

To improve generation quality while maintaining efficiency, viewpoint-conditioned multi-view diffusion has emerged as a key technique. These models synthesize consistent multi-view images conditioned on text or single-view inputs, which are then lifted into 3D through reconstruction. Zero123++~\cite{shi2023zero123++} models joint distributions of canonical views by tiling them into a single frame, while SyncDreamer~\cite{liu2024syncdreamer} enforces cross-view coherence via synchronized noise prediction and intermediate 3D feature volumes. Wonder3D~\cite{long2024wonder3d} employs cross-domain diffusion to generate multi-view normals and colors, followed by explicit geometry extraction through normal fusion. Era3D~\cite{li2024era3d} improves scalability and camera prior alignment with diffusion-based camera regression and efficient epipolar attention. Unique3D~\cite{wu2024unique3d} further advances multi-view generation and upscaling with a multi-stage surface reconstruction pipeline. 
% 
% More related work on lifting and reconstruction can be found in~\cite{cao2024supernormal,zhu2024pcf,zhu2025rethinking,zhu2026cos3d,zhu2026eps3d,wang2025vggt}. 
% 

Beyond 2D lifting, another promising direction is the development of native 3D diffusion and large-scale generative models that learn distributions directly over 3D latents or geometries. Direct3D~\cite{wu2024direct3d} introduces a triplane-latent VAE and diffusion transformer for image-conditioned generation without SDS. CLAY~\cite{zhang2024clay} scales to 1.5B parameters for geometry and PBR textures, while industrial systems such as Hunyuan3D 2.0~\cite{zhao2025hunyuan3d} integrate flow-based geometry diffusion with texture synthesis and production-ready editing tools. Large-scale pipelines like TRELLIS~\cite{xiang2025structured} and TRELLIS.2~\cite{xiang2025native} further advance scalability by combining hierarchical latent spaces with multi-resolution diffusion, enabling high-fidelity asset generation for professional content creation, with TRELLIS.2 supporting up to $1536^3$ through training-free upscaling. Representation-centric works also aim for high-resolution outputs. Specifically, recent work XCube~\cite{ren2024xcube} supports sparse voxel grids up to $1024^3$, and TetraDiffusion~\cite{kalischek2024tetradiffusion} uses tetrahedral partitions for efficient mesh generation.

Very recently, a concurrent work (\ie, LATTICE~\cite{lai2025lattice}) introduced VoxSet for high-resolution decoding, achieving increased resolution by training on a large-scale dataset at a substantial computational cost.
However, most approaches still require multi-scale high-resolution supervision or heavy multi-GPU training, motivating cost-effective test-time resolution scaling such as localized patch refinement that reuses object-level priors without \markhl{multi-scale} retraining.

\vspace{-1mm}
\subsection{Super-Resolution for 3D Content/Geometry}
\vspace{-1mm}
Rather than synthesizing textured meshes from scratch, another line of work enhances geometric surfaces of existing 3D models. Analogous to image super-resolution, point-cloud upsampling aims to densify sparse inputs uniformly on the underlying surface~\cite{yu2018pu,qian2020pugeo,qian2021deep,mao2022pu}, but generally does not introduce novel fine-scale geometry. Exemplar-based detail transfer can map local patterns from a high-quality source mesh to a target~\cite{berkiten2017learning}, and recent work extends such paradigms to coarse voxel refinement for few-shot detailization~\cite{chen2024decollage}. Text-guided mesh editing frameworks enable semantic manipulation but struggle with precise fine-grained control~\cite{gao2023textdeformer}. PBR material generation often estimates normal/bump maps that can be exploited as refinement signals~\cite{wang2024boosting}.

Diffusion-based SR approaches aim to recover high-frequency details directly. SDF-Diffusion~\cite{shim2023diffusion} adopts a two-stage pipeline, first generating a low-resolution signed distance field (SDF) and then applying a patch-based SR model trained on paired 
% high- and low-resolution 
samples. For radiance-field or Gaussian-based scenes, 3DSR~\cite{chen2025bridging} leverages off-the-shelf 2D diffusion SR within a 3D Gaussian-splat representation to enhance resolution while maintaining cross-view consistency. Mesh-oriented pipelines, e.g. SuperCarver~\cite{zhang2025supercarver}, refine multi-view normal maps via diffusion and update geometry through inverse rendering, achieving up to $16\times$ surface-detail enhancement with texture alignment. However, these methods either require high-resolution supervision or are tailored to mesh or radiance-field settings.

Existing SR methods improve fidelity but typically depend on high-resolution training or specialized architectures. In contrast, we propose a novel localized SR framework, providing a scalable solution for high-resolution detail under limited computational resources.

\begin{figure*}[t!]
    \centering
    % \vspace{2in}
    \includegraphics[width=\linewidth]{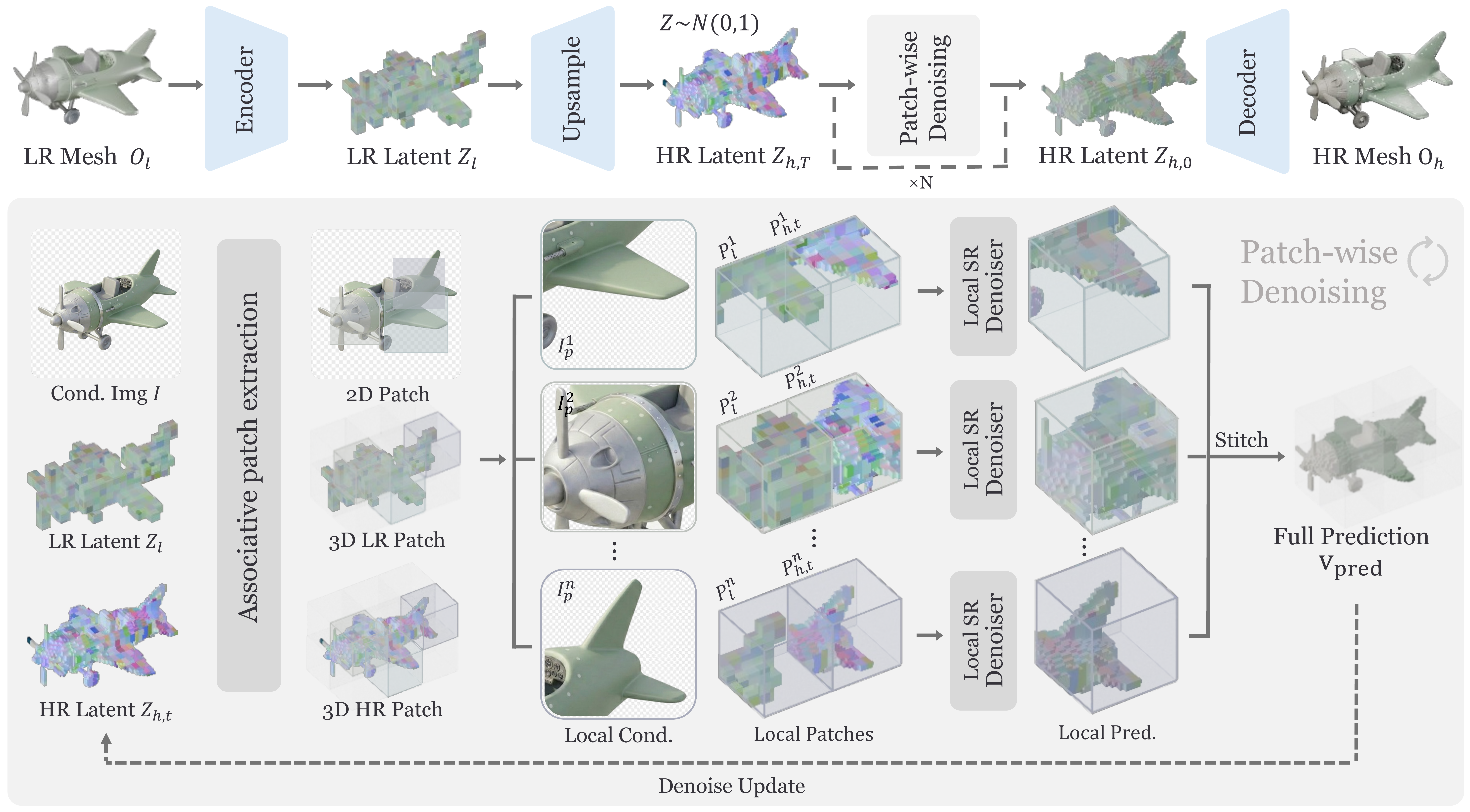}
    \caption{Overview of our PLSR. Starting from a coarse mesh $O_l$, we encode to obtain LR condition $Z_l$, upsample to obtain the sparse voxel skeleton of HR latent, and initialize with Gaussian noise to form $Z_{h,T}$. Rather than denoising $Z_h$ globally, we split $Z_h$ into fixed-size patches and crop $Z_l$ and image $I$ associatively to obtain a triplet, $(P_{h,t},P_l, I_p)$. 
    During iterative patch-wise denoising, our model iterates through the input tuples and predicts for each individual patch. The per-patch outputs are aggregated into a global prediction $v_{\rm pred}$ for denoising update. Once $Z_h$ is fully denoised, it can be decoded to an output mesh or be used as new LR condition $Z_l$ for the next SR round.
    Particularly, our localized SR process can be performed progressively to achieve higher resolution. 
    }
    \label{fig:overview}
\end{figure*}

\section{Method} 
\vspace{-1mm}

\label{sec:method}

\Cref{fig:overview} illustrates our \textbf{progressive} and \textbf{localized} super-resolution pipeline. Our goal is to generate a high‑resolution textured mesh $\mathcal{O}_{h}$ from a coarse input mesh $\mathcal{O}_{l}$ conditioned on one or more reference images $I$ with known camera parameters in a memory-efficient way.
We first introduce the basic concepts in \cref{sec:preliminaries}, associative input decomposition in \cref{sec:decomposition}, our localized SR model in \cref{sec:model}, model training in \cref{sec:training}, and inference pipeline in \cref{sec:pipeline}.

\subsection{Preliminaries: Structured Latents and Flow-Based Generation}
\label{sec:preliminaries}

\subsubsection{Structured sparse latent.}
Following \cite{xiang2025native}, we represent a 3D object by a compact VAE latent $Z$ defined on a sparse voxel support $\mathcal{S}_r \subset \mathbb{Z}^3$ at resolution $r$. Each active voxel index in $\mathcal{S}_r$ holds a $d$-dimensional feature, yielding $Z \in \mathbb{R}^{|\mathcal{S}_r|\times d}$. A pretrained encoder/decoder maps between meshes $\mathcal{O}$ and such sparse latents. In practice, TRELLIS.2 converts meshes into high-resolution ($1024^3$) intermediate o-voxel representation, and encodes the o-voxel representation with a convolutional VAE encoder to provide $\times16$ compression of the voxel representation along each axis, into a compact structured latent. The \emph{structured latent} itself takes a negligible amount of memory even at high-resolution. Its voxel structure also naturally supports easy patch operations, and stitching via weighted average, which we leverage for localized SR.

\subsubsection{Flow model: velocity and sampling.}
We generate (or refine) a latent $Z$ on a prescribed support $\mathcal{S}_r$ using a flow model \cite{lipman2022flow, liu2022flow} that predicts a time-dependent \emph{velocity} field $v_\theta(Z_t, t \mid \mathcal{C})$, where $Z_t$ denotes the latent at time $t\!\in\![0,1]$ and $\mathcal{C}$ collects conditioning signals (e.g., the LR latent $Z_l$, image crops $I_p$). Intuitively, $v_\theta$ defines how to \emph{move} the current latent to become slightly closer to the target data at time $t$. Starting from Gaussian noise $Z_{1}\!\sim\!\mathcal{N}(0,I)$ on $\mathcal{S}_r$, we integrate the ODE
\begin{equation}
\frac{d Z_t}{dt} \;=\; v_\theta(Z_t, t \mid \mathcal{C}),
\label{eq:flow_ode_simple}
\end{equation}
from $t{=}1$ to $t{=}0$ to obtain a clean latent $Z_{0}$, which is then decoded to a mesh $\mathcal{O}$. In practice we use a discrete solver with step size $\Delta t$, yielding the update
\begin{equation}
Z_{t-\Delta t} \;=\; Z_{t} \;+\; \Delta t \,\cdot\, v_\theta(Z_{t}, t \mid \mathcal{C}),
\label{eq:discrete_update}
\end{equation}
where each step can be interpreted as a small \emph{denoising} move guided by the predicted velocity. In later sections, we instantiate $\mathcal{C}$ with $(Z_l, I)$ and apply Eq.~\eqref{eq:discrete_update} in a patch-wise manner. 

% \subsection{Localized Super-Resolution Framework}% 

\subsection{Associative Input Decomposition}
\label{sec:decomposition}
Different from a typical image-guided 3D super-resolution setup where all inputs are global, our localized super-resolution requires local input triplets (noisy high-resolution latent patch $P_h$, low-resolution latent patch $P_l$, and corresponding local reference image $I_p$) to perform image-conditioned local super-resolution. This aligns with the core idea of learning relative super-resolution independent of absolute object resolution, as it encourages the model to learn its super-resolution mostly on local information while preventing it from relying on global cues to form its outputs.

However, the local conditioning image has to be defined carefully and in a consistent way such that a similar image distribution would apply to both training and inference to minimize the gap. Considering users typically provide only global reference images as conditions in practice, the local conditioning image must be easily obtainable, given the global image. Therefore, we choose to use image crops from a global reference image centered around the target 3D patch as a suitable proxy of local reference image of the patch, as a crop can contain most necessary information to generate local details with minimal irrelevant information. 

To obtain the triplet, we associatively extract the patches to ensure their alignment. Given a patch bounding box in 3D space, we first locate and crop the partial structured latent $P_h$ and $P_l$ from the global latent grids. For patch image condition, we project the patch bounding box onto the conditioning image using known camera parameters, thereby extracting localized image features. By cropping the conditioning image with this patch bounding box, we effectively obtain a close-up shot of the patch currently being refined, providing both localized and fine-grained features necessary for refinement. This patch extraction method is used in both training and inference for consistency.

% \vspace{.05in}
% \noindent\textbf{Patch-Space Voxel Coordinate for Localized Positional-Awareness.}
As our base flow model embeds the positions of active voxels of the structured latent into features to ensure positional-awareness during generation, we choose to enforce both LR and HR patch to use voxel coordinates relative to the patch origin instead of the global coordinates in both training and inference, preventing the model from relying on global positional cues for prediction.
% \subsection{Localized Super-Resolution Architecture}

\begin{figure}[!t]
    \centering
    \includegraphics[width=0.85\linewidth]{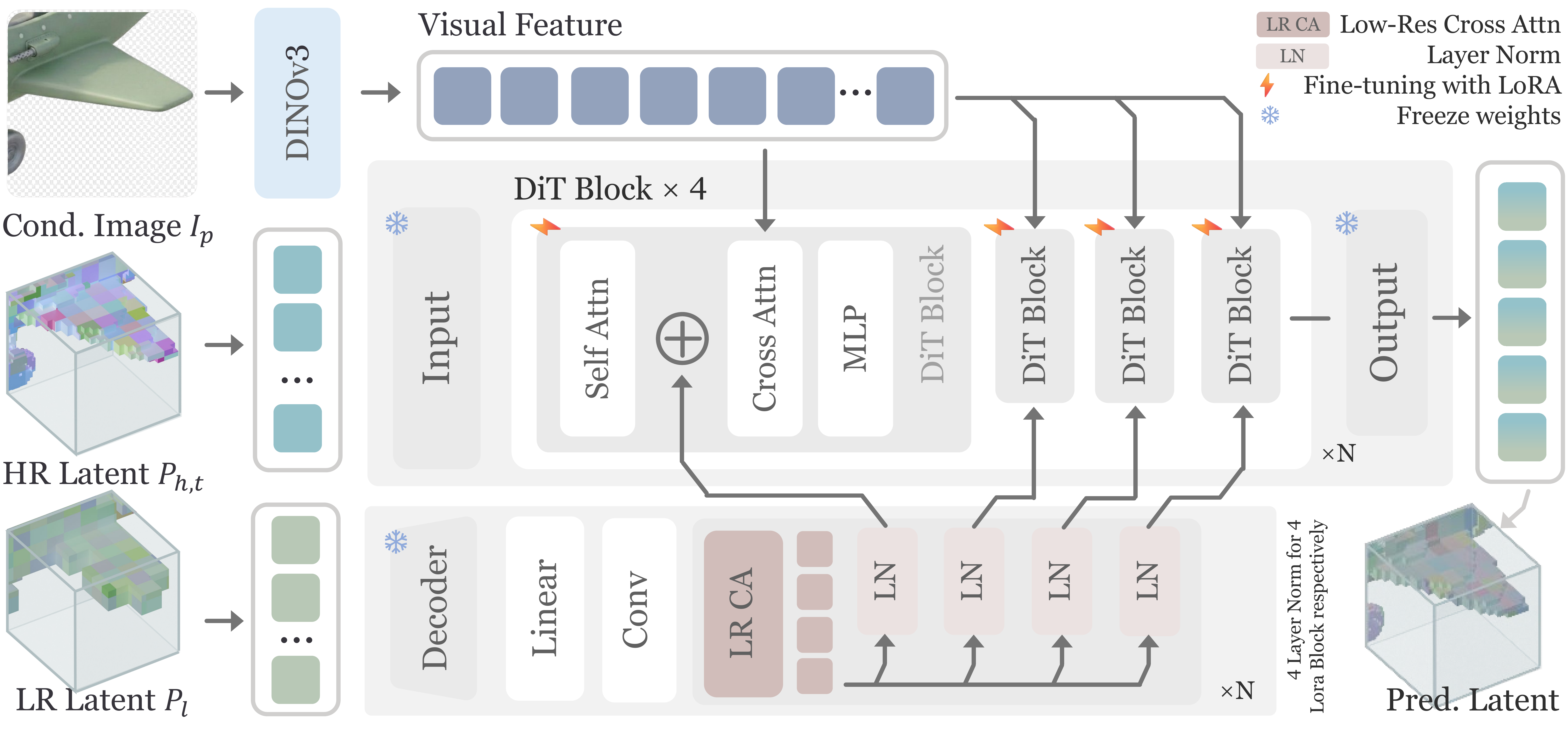}
    %\vspace{2in}
    \caption{Architecture of our proposed localized SR model.}
    \label{fig:architecture}
\end{figure}

\subsection{Localized Super‑Resolution Model}% (subsubsections for structure clarity?)}
\vspace{.05in}
\label{sec:model}
\noindent\textbf{Basic architecture.}
We base our model on the pretrained sparse latent DiT model proposed in TRELLIS.2~\cite{xiang2025native}. The original DiT model consists of a sequence of transformer blocks with image cross attention, AdaLN-single modulation and Rotary Position Embedding. Given noisy structured latent $Z_t$, the flow model predicts velocity $v$ conditioned on DINOv3 image features $F_I$ extracted from $I$ and timestep $t$. We incorporate an LR conditioning branch into the model to take additional LR conditioning features extracted from LR latent patch using the first two decoder blocks of the pretrained VAE decoder, denoted as  $\mathcal{D}_2$. The last layer of $\mathcal{D}_2$ upsamples the LR feature by $\times2$, aligning it with $P_{h,t}$. The model architecture is shown in \cref{fig:architecture}. Formally, our model $v_\theta$ predicts velocity:

\begin{equation}
    v = v_\theta(P_{h,t}, \mathcal{D}_2(P_l), F_{I_p},t),
\end{equation}
where $P_{h,t}$ denotes noisy high-resolution latent patch at time $t$; $P_l$ denotes its corresponding low-resolution latent patch; $I_p$ denotes the corresponding image patch obtained by randomly extracting patches from training data associatively.

\vspace{.05in}
\noindent\textbf{Lightweight LR Attention for Flexible Conditioning.}
Naturally, channel-wise concatenating the condition to input is a simple-yet-effective way to achieve spatially aligned conditioning. While our LR feature $\mathcal{D}_2(P_l)$ is at the same resolution as input $P_h$, we observed that they exhibit slight voxel coordinate discrepancies, probably due to slight shape changes that occur during remeshing and simplification used to degrade LR mesh. This small discrepancy makes channel concatenation strategies ineffective due to the strict spatial alignment requirement. We instead choose to introduce a lightweight LR attention layer to overcome the mismatch problem. These layers compute attention between each HR voxel and its neighboring LR cells within a small radius, enabling flexible information transfer without resorting to heavy global attention. 
To keep the LR conditioning branch lightweight, we include a dedicated LR attention layer only in a subset of transformer blocks of the original flow model, and reuse the resulting features across subsequent blocks until a new attention computation is performed. To ensure the compatibility with the feature statistics of each block, we apply layer normalization with element-wise affine transformation when reusing LR features. The LR features are added element-wise  to transformer feature after self-attention layer. In this way, the trainable parameter in our fine-tuning is substantially reduced, while providing necessary LR signal for robust SR.

\subsection{Model Training}
\label{sec:training}
To adapt the model to the new task, we adopt LoRA~\cite{hu2022lora} fine-tuning on all linear layers except the input and output layer, avoiding catastrophic forgetting or unstable optimization when fine-tuned on limited dataset and resources. To speed up learning, we reuse the weights from original self-attention layer to initialize our LR lightweight attention layer.

We train our model using standard Conditional Flow Matching objective:
\begin{equation}
    \mathcal{L}_{CFM}(\theta) = \mathbb{E}_{t,P_{h,0}, \epsilon}||v_\theta(P_h(t), \mathcal{D}_2(P_l), F_{I_p},t) - (\epsilon - P_{h,0})||^2_2
\end{equation}
 During training, $P_h(t) = (1-t)P_{h,0} + t\epsilon $ is randomly sampled by injecting Gaussian noise $\epsilon$ into $P_{h,0}$ based on random timestep $t$. A core benefit of this patch formulation lies in its memory and data efficiency during training. On one hand, encoding meshes at high-resolution (i.e., $96^3$ latent resolution and beyond) is extremely resource-intensive or even infeasible due to GPU memory constraints. On the other hand, high-quality datasets are not widely accessible. We instead train on abundant $P_{h}$ data that can be diversely sampled from moderate-resolution data, bypassing both limitations. To be specific, we encode textured meshes at the same global resolution as in TRELLIS.2~\cite{xiang2025native} ($64^3$ latent resolution) and define our model's output patch size to be $32^3$, covering half of the volume along each axis.

\subsection{Patch-wise Super-Resolution Pipeline}
\label{sec:pipeline}
Our pipeline performs patch-based super-resolution in the latent domain, leveraging its compactness and smoothness to keep the full latent representation in memory. The voxel space is partitioned into local patches, which are processed sequentially and later stitched together via weighted feature aggregation.  
Benefiting from the iterative denoising nature of flow models, we place our split-refine-and-stitch process inside each denoising iteration, so that a consensus prediction will be formed by weighted aggregation of per-patch predictions, enabling mutual awareness of neighboring patches, and providing a unified basis during the next denoising iteration. Compared to fully independent processing and one-time stitching over the output, this iterative aggregation yields better coherence, aligning with the observations in earlier works on image and texture synthesis\cite{bar2023multidiffusion, liu2024text}.

% \vspace{.05in}
\subsubsection{Initialization and Conditioning.}
The pipeline begins with a low-resolution sparse latent (half the target resolution), obtained either from coarse mesh encoding or from previous SR outputs. We first use the pretrained decoder to upsample the LR condition into sparse features at the same resolution as the target output to perform resolution-aligned conditioning. 
 Due to the fact that compressed latents operate at $1/16$ resolution per-axis to the raw voxel representation, it is in most cases safe to assume that the voxel skeleton extracted from the coarse object shares an indistinguishable structure compared to its HR counterpart. Thus, for inference, we could reuse the above-mentioned voxel indices to initialize noisy HR latent for denoising.

% \vspace{.05in}
\subsubsection{Patch Processing and Image Conditioning.}
At each denoising iteration, the 3D space is partitioned into equal-sized windows that cover the entire latent grid, with partial overlaps for neighbor aggregation, and we use associative patch extraction to obtain triplets for each window. 
When multiple conditioning images are provided by the user, we employ a visibility-based strategy to select one informative image for each denoising step. Specifically, we render the patch voxel structure with a voxel renderer from each camera, and apply depth testing to discard pixels where the target patch is occluded by surrounding regions, yielding an occlusion-aware visibility score for each image. Rather than always choosing the image with maximum visibility, we sample images weighted by visibility at each denoising step, thereby balancing local relevance with global  information coverage. For patches occluded from all views, the DINOv3 image features are zeroed-out to indicate no visual cues, falling back to LR 3D latents for stability.

% \vspace{.05in}
\subsubsection{Aggregation and Global Coherence.}
After predicting per-patch velocity with the flow model, all patches are aggregated at their overlapping region by weighted averaging. As patch-based predictions tend to be less reliable at patch boundaries due to an incomplete context, we weigh patch higher near the center and lower at their boundaries during merging. The aggregated global velocity is used to update the denoising process, i.e., to compute $Z_{h,t-1}$ at the next time step. This $Z_{h,t-1}$ then serves as the input noisy latent in the subsequent iteration, ensuring that patch-wise processing in the next iteration starts from a unified input state.

% \vspace{.05in}
\subsubsection{Progressive Super-Resolution.}
Our method supports multi-round SR in latent space by feeding the output latent into the next round of SR as low-resolution latent. Each round of super-resolution increases the resolution by a factor of two along each axis, and super-resolving for multiple rounds supports $\times4$ or more super-resolution relative to the initial LR resolution. 

\subsubsection{Practical Workflow.}
\markhl{In practical image-to-3D workflows where a reference image is present, we can invoke an existing 3D generator to obtain the coarse mesh, align the user-provided reference image with the mesh, and refine based on the estimated camera parameter to boost resolution. For mesh-only input scenarios, users may render the object from known camera views and obtain detailed conditioning images through image editing tools, manual painting, or both.}

% \vspace{.05in}
\subsubsection{Selective Super-Resolution.}
Due to the localized nature of our model, it can also be used for selective super-resolution to refine only in user-specified regions for efficiency. However, due to page limitation, we defer the discussion of this application to the supplementary materials.

\section{Experiments}
\label{sec:experiments}
\subsection{Experimental Setting}
\label{sec:implementation}
\subsubsection{Implementation Details.}  
We train on 30K assets from Objaverse‑XL~\cite{deitke2023objaversexl}, filtered by aesthetic score~\cite{schuhmann_improved-aesthetic-predictor_2026} $\ge6.5$. Each asset is encoded into geometry and texture latents using the TRELLIS.2 VAE at a latent resolution of $64^3$ (corresponding to $1024^3$ o‑voxels~\cite{xiang2025native}). A low‑resolution version is obtained by simplifying and re‑encoding the same asset at $32^3$ resolution. We render 24 viewpoints per asset in Blender 3.3 LTS~\cite{blender33lts} with randomized FOV settings at $2048^2$ resolution to incorporate fine visual details.
We sample 3D patches using our associative extraction strategy, producing HR/LR latent pairs of size $32^3$/$16^3$ along with local image crops at $1024^2$ resolution. All parameters of the pretrained \markhl{ TRELLIS.2-4B shape and texture flow DiT } are frozen, and we insert LoRA~\cite{hu2022lora} adapters into all linear layers except the input/output layers. Geometry and texture models are trained using flow‑matching loss with classifier‑free guidance (drop rate 0.3). A light regularization loss ($\lambda$ = 0.005) discourages trivial opening in the image-modulation path.
Optimization uses AdamW (lr = $1\times10^{-5}$, weight decay = 0.1) with EMA (0.999). Random color jitter, geometric warp, and local dropout augmentations simulate inconsistencies in patch‑aligned conditioning images. Both geometry and texture SR models are trained for 2 epochs, with 2 objects per batch and up to 3 patches per object, requiring approximately 50 GPU‑hours per model.
During inference, we encode the coarse object at initial resolution $640^3$ and progressively super-resolve it to resolutions of $2560^3$ by conducting twice the SR operation.
Besides, to use a user-specified image as the condition for inference, our method uses \markhl{MapAnything}~\cite{keetha2025mapanything} to predict the camera parameters relative to the coarse object. 
All experiments are conducted on a single NVIDIA A100 GPU with 80 GB memory and implemented using PyTorch. Refer to the supplementary material for runtime and memory on average use case.

\subsubsection{Datasets and Evaluation Metrics.}
\markhl{We evaluate our method under two complementary evaluation settings: 
a pseudo-ground-truth test set for objective metrics, and an open-world challenging test set for VLM and human  evaluation.}

\paragraph{Pseudo-GT evaluation.}
\markhl{Following the evaluation protocol of TRELLIS.2, we collect around 90 complex and detail-rich 3D assets from the Sketchfab staff-picked} list~\cite{sketchfab_staffpicks} \markhl{as pseudo ground truth. For each asset, we render the original high-resolution object as the image condition, and construct the low-resolution input by remeshing the geometry and baking low-resolution textures in }Blender 5.1.1~\cite{blender511}, \markhl{removing high-frequency geometry and texture details. This setting allows us to evaluate generative refinement ability using both 2D render-space and 3D geometry-space metrics. For render-space perceptual similarity, realism, and detail richness, we render each generated asset from four preset views and compute} LPIPS~\cite{zhang2018unreasonable}, FID~\cite{heusel2017gans}, and the High Frequency Ratio (HFR), defined as
$
\mathrm{HFR}(X)=
\frac{\mathrm{mean}(|X-\mathrm{Gaussian}(X)|)}
{\mathrm{mean}(|X|)+\epsilon}.
$
\markhl{For 3D geometric fidelity, we report Chamfer Distance (CD) and F1 Score in normalized space with a threshold of $10^{-6}$. Point clouds for 3D metrics are sampled only from visible outer surfaces, obtained by unprojecting depth maps, with 1M points per asset.}

\paragraph{Open-world evaluation.}
We further assemble an open-world test set containing around 70 challenging objects, \eg, characters, animals, mechanical assets, props, and miniature scenes. Each sample includes a reference image depicting a challenging object and a low-resolution textured mesh generated by a pretrained 3D generator~\cite{xiang2025native,zhao2025hunyuan3d}. Since no ground-truth high-resolution mesh is available in this setting, following GPTEval3D~\cite{wu2024gpt}, we adopt a VLM-based pairwise evaluation using GPT-5.2~\cite{openai_gpt52}. The VLM refers to the reference image and a pair of method outputs rendered at $1024^2$ resolution, using normal renders for geometry evaluation and colored renders for texture evaluation. \markhl{It compares the outputs along four aspects: detail richness (\textit{Detail Rich}), alignment with the reference image (\textit{Align w/ Ref.}), artifact-freeness (\textit{Artifact Free}), and overall quality (\textit{Overall Quality}). Each pairwise comparison is repeated three times to reduce stochasticity, and final rankings are computed using an }ELO~\cite{elo1967proposed} estimator. \markhl{Details on data list/identifiers, VLM prompts and  additional human evaluations on the same dimensions are included in the supplementary materials.}

\subsubsection{Baselines.}
We compare with TRELLIS.2~\cite{xiang2025native}, UltraShape~\cite{jia2025ultrashape}, DetailGen3D~\cite{deng2026detailgen3d}, and MVPaint UVR~\cite{cheng2025mvpaint}. \markhl{SuperCarver~\cite{zhang2025supercarver} and LATTICE~\cite{lai2025lattice} are not included because their official code/checkpoints are unavailable, making controlled comparisons infeasible. We instead include UltraShape as an open-source implementation of LATTICE for geometry refinement. For TRELLIS.2, we use its self-cascading refinement pipeline and evaluate it at both its default resolution ($1536^3$) and our output resolution ($2560^3$). Both TRELLIS.2 and UltraShape initialize their voxel latent support from the input coarse mesh. DetailGen3D serves as an image-guided geometry detailization baseline, while MVPaint UVR is used for texture refinement. Geometry- or texture-only baselines are evaluated only on applicable metrics.}

\subsection{Results}
\label{sec:results}

\subsubsection{Qualitative Comparison.}
\label{sec:qualitative}
We provide the visual comparisons between our method and two of the most competitive baselines (\ie, TRELLIS.2 and UltraShape) in \cref{fig:qualitative}. Remaining baselines are compared in the supplementary materials. 
The results clearly demonstrate that our method consistently reconstructs sharper fine‑grained geometry and richer surface appearance, while maintaining global coherence and avoiding patch seams.
\begin{figure*}[!t]
    \includegraphics[width=\linewidth]
    {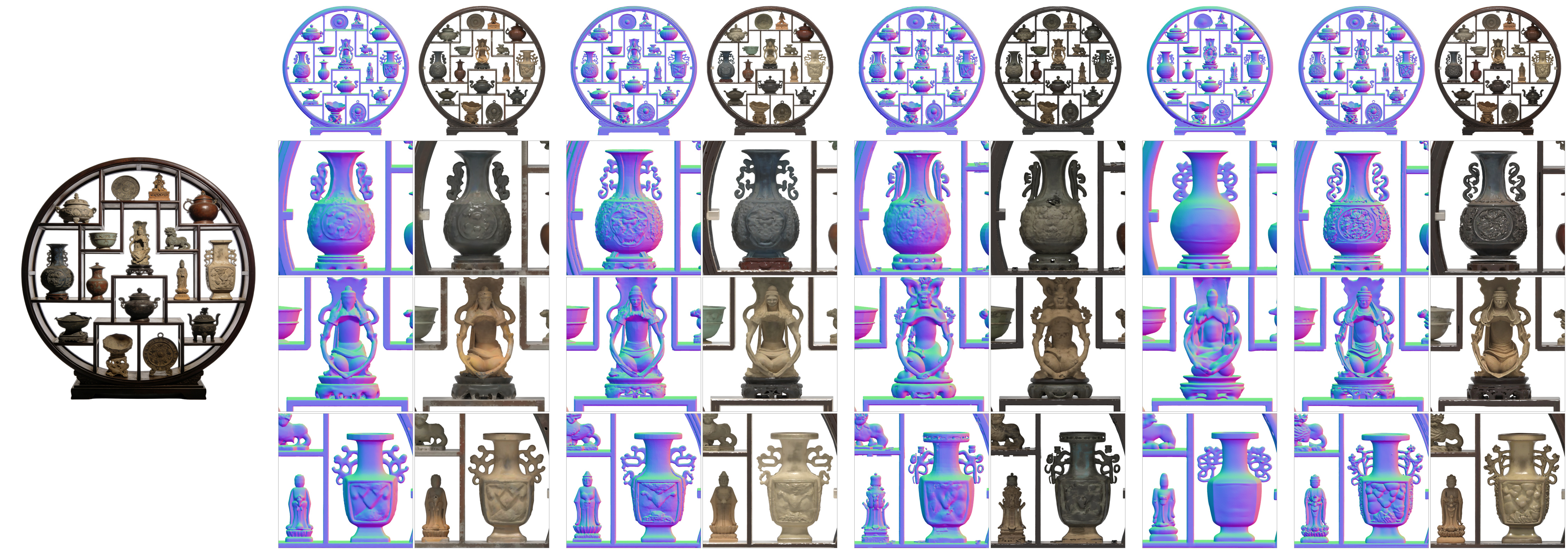}\par
    \vspace{1ex}
    \includegraphics[width=\linewidth]
    {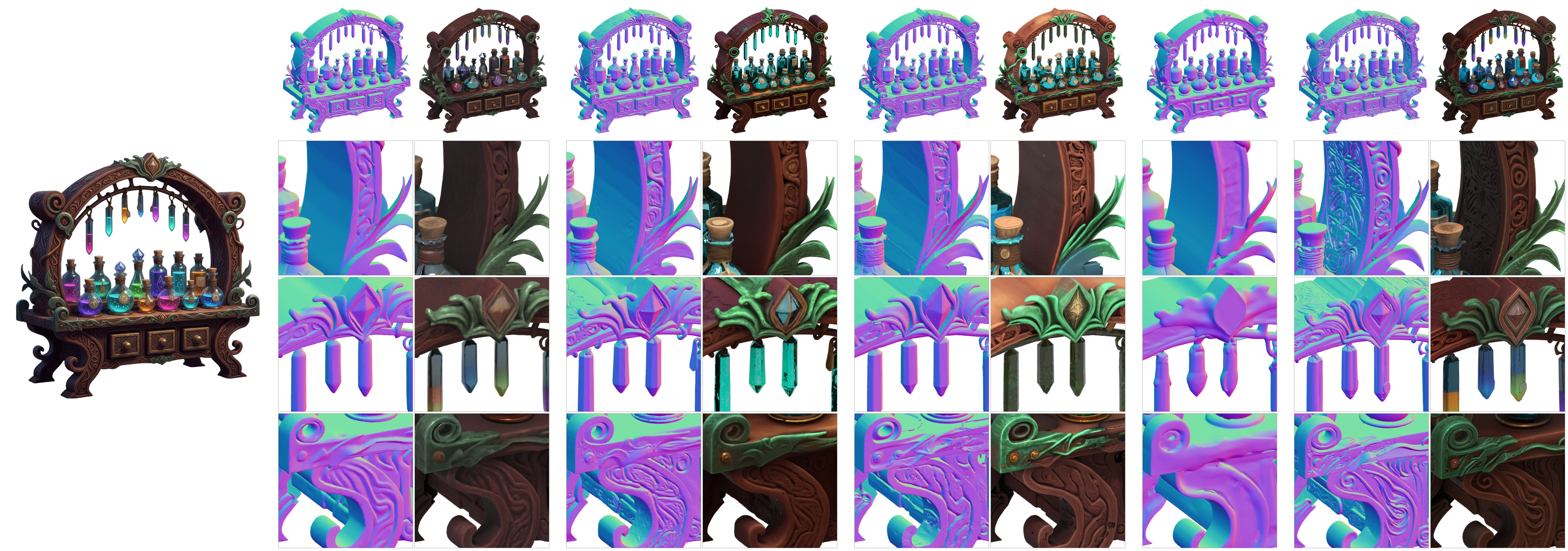}\par
    \vspace{1ex}
    \includegraphics[width=\linewidth]
    {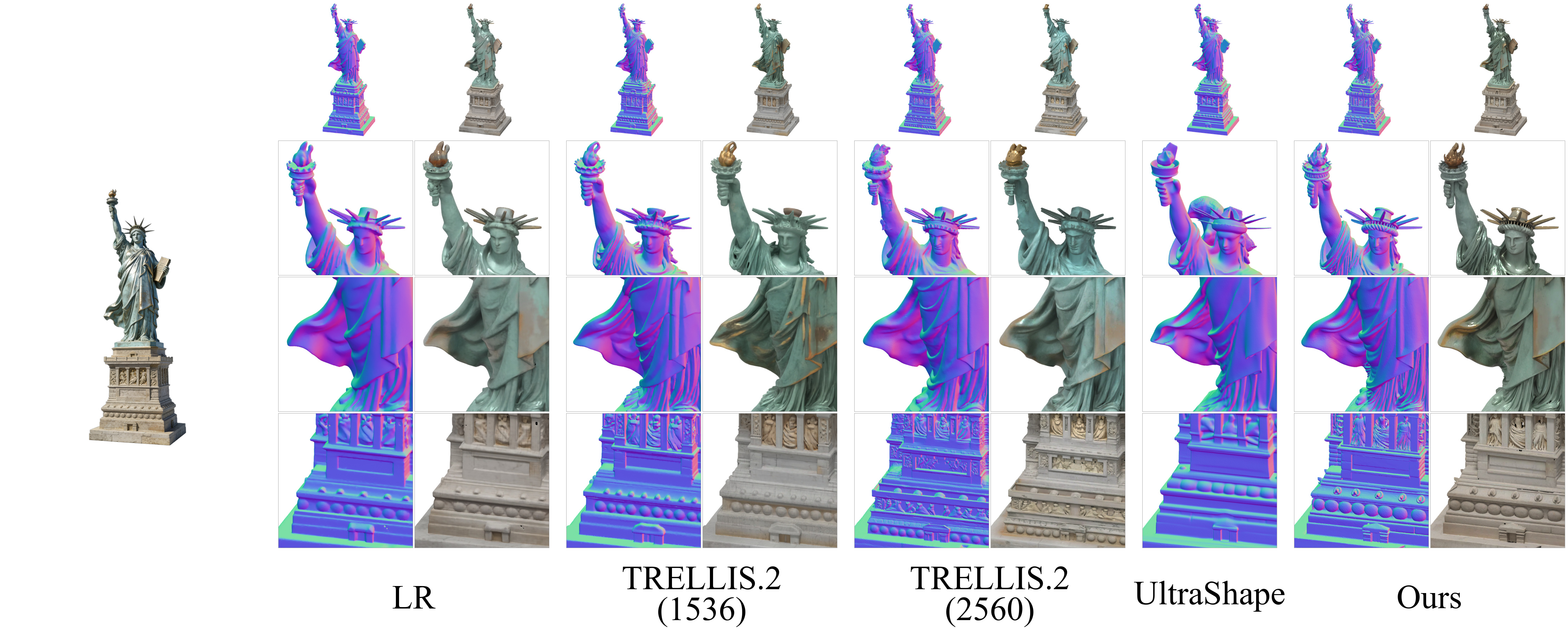}\par
    \caption{\markhl {Qualitative comparison on complex shape. Our method recovers fine details and avoids seams, while baselines show limited detail refinement ability. Zoom-in recommended.}}
    \label{fig:qualitative}
    % \vspace{-0.5em}
\end{figure*}

\newcolumntype{Y}{>{\centering\arraybackslash}X}
\begin{table}[t!]
    \centering
    \caption{\markhl {Objective pseudo-GT evaluation. CD/HFR are scaled as indicated.}}
    \label{tab:pseudogt}
    \scriptsize
    \setlength{\tabcolsep}{2.5pt}
    \renewcommand{\arraystretch}{1.05}

    \begin{minipage}[t]{0.57\linewidth}
    \vspace{0pt}
    \centering
    \textbf{Obj. Geometry Metrics} {(CD$\times10^3$, HFR$\times10^2$)}\\[-0.2ex]
    \begin{tabularx}{\linewidth}{l*{5}{Y}}
        \toprule
        {Method } &
        {LPIPS$_\downarrow$} &
        {FID$_\downarrow$} &
        {HFR$_\uparrow$} &
        {CD$_{\downarrow}$} &
        {F1$_\uparrow$} \\
        \midrule
        TR.2(1536) & \underline{0.122} & \underline{60.37} & \underline{0.981} & 0.0247 & 0.059\\
        TR.2(2560) & 0.122 & 63.19 & 0.923 & \underline{0.0225} & 0.036 \\
        UltraShape      & 0.167 & 72.03 & 0.876 & 1.3982 & \underline{0.109} \\
        DetailGen3D     & 0.187  & 93.35  & 0.645 & 0.4600 & 0.083  \\
        Ours(2560)      & \textbf{0.109} & \textbf{55.89} & \textbf{1.088} & \textbf{0.0139} & \textbf{0.145} \\
        \bottomrule
    \end{tabularx}
    \end{minipage}
    \hfill
    \begin{minipage}[t]{0.41\linewidth}
    \vspace{0pt}
    \centering
    \textbf{Obj. Texture Metrics} (HFR$\times10^2$)\\[-0.2ex]
    \begin{tabularx}{\linewidth}{l*{3}{Y}}
        \toprule
        {Method } &
        {LPIPS$_\downarrow$} &
        {FID$_\downarrow$} &
        {HFR$_\uparrow$} \\
        \midrule
        TR.2(1536) & \underline{0.150} & \underline{70.20} & \underline{1.501}  \\
        TR.2(2560) & 0.158              &  72.77            & 1.482  \\
        MVPaint         & 0.154              &  77.18            & 1.052\\
        Ours(2560)      & \textbf{0.125}    & \textbf{59.89}    & \textbf{1.646} \\
        \bottomrule
    \end{tabularx}
    \end{minipage}
\end{table}

\begin{table}[t!]
    \centering
    \caption{\markhl {Open-world evaluation. ELO}~\cite{elo1967proposed} \markhl {ratings for geometry and texture performance are reported in separate tables for clarity.}}
    \label{tab:gptmetrics}
    \scriptsize
    \setlength{\tabcolsep}{2.5pt}
    \renewcommand{\arraystretch}{1.05}

    \begin{minipage}[t]{0.49\linewidth}
    \vspace{0pt}
    \centering
    \textbf{Geometry ELO $\uparrow$}\\[-0.2ex]
    \begin{tabularx}{\linewidth}{l*{4}{Y}}
        \toprule
        \makecell{Method} &
        \tiny \makecell{Detail\\Rich} &
        \tiny \makecell{Align\\w/ Ref.} &
        \tiny \makecell{Artifact\\Free} &
        \tiny \makecell{\textbf{Overall}\\\textbf{Quality}} \\
        \midrule
        TR.2(1536) & \underline{1775} & \underline{1774} & \textbf{1736} & \underline{1790} \\
        TR.2(2560) & 1627 & 1592 & 1492 & 1572 \\
        UltraShape      & 1246 & 1385 & 1551 & 1356 \\
        DetailGen3D     & 869  & 927  & 1111 & 925  \\
        Ours(2560)      & \textbf{1982} & \textbf{1821} & \underline{1608} & \textbf{1855} \\
        \bottomrule
    \end{tabularx}
    \end{minipage}
    \hfill
    \begin{minipage}[t]{0.49\linewidth}
    \vspace{0pt}
    \centering
    \textbf{Texture ELO $\uparrow$}\\[-0.2ex]
    \begin{tabularx}{\linewidth}{l*{4}{Y}}
        \toprule
        \makecell{Method} &
        \tiny \makecell{Detail\\Rich} &
        \tiny \makecell{Align\\w/ Ref.} &
        \tiny \makecell{Artifact\\Free} &
        \tiny \makecell{\textbf{Overall}\\\textbf{Quality}} \\
        \midrule
        TR.2(1536) & \underline{1553} & \textbf{1621} & \textbf{1624} & \underline{1599} \\
        TR.2(2560) & 1494 & 1426 & 1441 & 1416 \\
        MVPaint         & 1197 & 1369 & 1358 & 1350 \\
        Ours(2560)      & \textbf{1756} & \underline{1582} & \underline{1575} & \textbf{1633} \\
        \bottomrule
    \end{tabularx}
    \end{minipage}
    \vspace{-1em}
\end{table}

\subsubsection{Quantitative Comparison.}
\label{sec:quantitative}
On the pseudo-GT test set (\cref{tab:pseudogt}), \markhl{our method achieves the best results across all reported geometry and texture metrics. The higher HFR, together with lower LPIPS and FID, suggests that our method recovers perceptually meaningful high-frequency details rather than merely amplifying noise. The lower CD and higher F1 further indicate better fidelity to the pseudo-GT geometry. }

\Cref{tab:gptmetrics} further reports VLM-based ELO ratings on the open-world test set. Our method achieves the highest scores in detail richness, image alignment, and overall quality for geometry refinement, and in detail richness and overall quality for texture refinement. It also ranks second on the remaining criteria, including artifact-freeness and texture alignment. These results show that our local super-resolution not only scales textured meshes to a substantially higher resolution, \ie, $\times 2.5$ of the TRELLIS.2 base resolution, but also does so in a perceptually meaningful and image-faithful way. 

We also observe that directly increasing the TRELLIS.2 resolution to $2560^3$ leads to degraded performance, likely due to limited generalization beyond its training resolution, further highlighting the value of progressive local super-resolution for extreme-resolution refinement.

\begin{figure}[!t]
    \centering
    \includegraphics[width=0.9\linewidth]{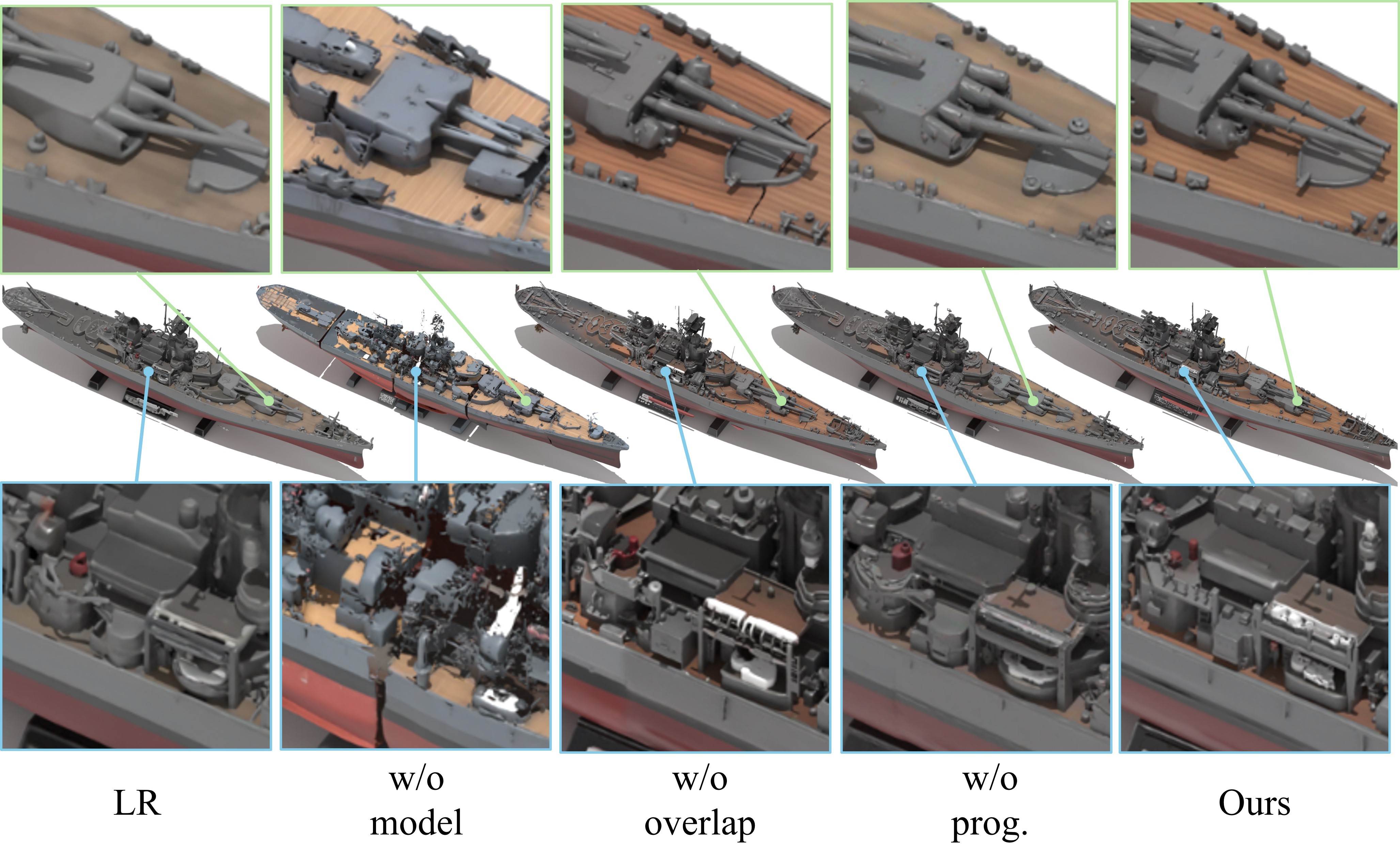}
    \caption{\markhl{Qualitative ablation study.}
    Given LR cues as anchors, our trained model suppresses most seams
even without overlaps, and with overlapping applied, the seams are fully removed. }
\label{fig:ablationfigure}
% \vspace{em}
\end{figure}

\begin{table}[t!]
    \centering
    \caption{\markhl {Quantitative ablation study.
    We compare our method in a patch-based inference setting with different alternatives (\ie,
    TRELLIS.2 base model with our pipeline (w/o model), our method without overlapped inference (w/o overlap), our method without progressive SR (w/o prog.).} }
    \label{tab:quantitative_ablation}
    \scriptsize
    \setlength{\tabcolsep}{2.5pt}
    \renewcommand{\arraystretch}{1.05}

    \begin{minipage}[t]{0.49\linewidth}
    \vspace{0pt}
    \centering
    \textbf{Ablation Geometry ELO $\uparrow$}\\[-0.2ex]
    \begin{tabularx}{\linewidth}{l*{4}{Y}}
        \toprule
        \makecell{Method} &
        \tiny \makecell{Detail\\Rich} &
        \tiny \makecell{Align\\w/ Ref.} &
        \tiny \makecell{Artifact\\Free} &
        \tiny \makecell{\textbf{Overall}\\\textbf{Quality}} \\
        \midrule
        w/o model & 1291 & 1281 & 1209 & 1281 \\
        w/o overlap &\underline{ 1611} & \underline{1559} & 1550 & \underline{1574} \\
        w/o prog.      & 1424 & 1552 & \textbf{1665} & 1537 \\
        Ours      & \textbf{1672} & \textbf{1606} & \underline{1575} & \textbf{1605} \\
        \bottomrule
    \end{tabularx}
    \end{minipage}
    \hfill
    \begin{minipage}[t]{0.49\linewidth}
    \vspace{0pt}
    \centering
    \textbf{Ablation Texture ELO $\uparrow$}\\[-0.2ex]
    \begin{tabularx}{\linewidth}{l*{4}{Y}}
        \toprule
        \makecell{Method} &
        \tiny \makecell{Detail\\Rich} &
        \tiny \makecell{Align\\w/ Ref.} &
        \tiny \makecell{Artifact\\Free} &
        \tiny \makecell{\textbf{Overall}\\\textbf{Quality}} \\
        \midrule
        w/o model & 1288 & 1270 & 1269 & 1271 \\
        w/o overlap  & \underline{1624} & \underline{1606} & 1537 & \underline{1609}\\
         w/o prog.       &  1441 & 1509 & \underline{1595} & 1475 \\
         Ours      &  \textbf{1645} & \textbf{1613} & \textbf{1597} & \textbf{1643} \\
        \bottomrule
    \end{tabularx}
    \end{minipage}
    \vspace{-1em}
\end{table}

\subsection{Ablation Study}
\label{sec:ablation}
We conduct ablations \markhl{via VLM-based ratings} to verify the effectiveness of each component by comparing our full method with three variants. The first variant uses the original TRELLIS.2 flow model without our modification and training for local SR prediction (w/o model). The second variant conducts single-stage refinement without multi-round SR (w/o progressive). 
The third removes overlaps between local patches to evaluate the effectiveness of our synchronized denoising pipeline (w/o overlap). Both the qualitative results in \cref{fig:ablationfigure} and quantitative results in \cref{tab:quantitative_ablation} demonstrate that our local SR model achieved robust refinement based on LR anchors and inference-time overlapping further suppresses seam artifacts. 
Moreover, non-progressive refinement achieved limited detailization, showing that the progressive refinement helps enhance multi-scale details. 

\paragraph{Effect of conditioning views.} 
\markhl{We further evaluate the effect of using multiple conditioning views on the pseudo-GT test set. Increasing the number of conditioning views from one to four improves normal/color LPIPS from 0.109/0.125 to 0.103/0.119, showing that additional views provide better coverage and reduce ambiguity in local refinement. Together with the qualitative ablation in the supplement, these results suggest that reference images provide useful instance-specific cues complementary to the LR 3D latent, while regions with limited visibility still rely more on the LR latent condition.

}

\section{Limitation}
Our method relies on overall accurate camera parameters and conditioning image quality. 
Severe misalignment or unrealistic conditioning images may introduce inconsistencies. Additionally, the coarse mesh must be structurally adequate, since our SR model is not designed to correct large geometric errors. Also, transfer to a different sparse voxel latent representation requires retraining the adapter.
% We leave these for future work.

\section{Conclusion}
\label{sec:conclusion}
We presented a resource-efficient, progressive and localized SR framework for high-resolution 3D asset generation without high-resolution retraining or multi-GPU compute. \markhl{By using structured sparse latents as a compact intermediate representation, PLSR decomposes the global mesh super-resolution into local conditional denoising tasks and progressively synthesizes fine geometry and texture details.}
Experiments demonstrate consistent improvements \markhl{over strong baselines in geometric and texture detail, suggesting that localized latent SR is a practical direction for scalable high-resolution 3D asset generation and refinement.}

{\small
\section*{{Acknowledgements.}}
\markhl{This research was supported in part by the Australian Government through the Australian Research Council (ARC) Discovery Project DP260100218.}
}

\bibliographystyle{splncs04}
\bibliography{main}

\clearpage


\section*{Supplementary material}

# A Selective Refinement

Our selective refinement procedure follows the same progressive super-resolution pipeline described in Sec. 3.5 of the main paper, but it restricts refinement to a user-specified spatial region instead of the full object. Starting from an axis-aligned bounding box $B$ provided by the user, we expand it by a padding of $p$ voxels on each side to form an enlarged region $\tilde{B}$. This padded region provides the necessary structural context so that any high-resolution details synthesized inside $B$ maintain a natural and coherent connection with their surroundings.

To begin refinement, we encode the coarse input object at the target resolution, but only within $\tilde{B}$. This produces a partial structured latent $Z_0$ that serves as the context for the selective SR process. During each denoising iteration, we execute the same patch-wise refinement strategy used in the global case, but we extract patches only within $\tilde{B}$. For every extracted patch, the localized SR model predicts a velocity, and all predictions are then aggregated to produce a single denoising update over the partial latent.

After each aggregation step, we apply a masked substitution procedure adapted from blended diffusion [1]. The newly refined latent values are preserved inside the target region $B$. In contrast, the padded band $\tilde{B} \setminus B$ is overwritten using a time-aligned noisy version of the partial latent $Z_0$. This strategy keeps the padded region consistent with the original coarse input throughout the denoising process and prevents artifacts from emerging along the refinement boundary.

The same procedure naturally extends to multi-round progressive super-resolution. Before each new round, we re-encode the partial object so that its latent representation matches the next-round resolution. After all rounds are completed, the refined region is composited back into the original coarse object to obtain the final output.

As shown in Fig. 1, even though refinement is restricted to localized areas, the resulting high-resolution details integrate naturally with the unrefined regions. The computation is also significantly reduced. For example, the three demonstrated cases achieved per-iteration speed-ups of 53.67% (hunter), 72.86% (town), and 55.26% (robot) when compared to global refinement using TRELLIS.2 baseline at the same $2560^3$ resolution. This makes selective refinement particularly beneficial when working with large objects or assets where only specific regions require additional detail.

# B User Study

To complement the VLM-based evaluation described in Sec. 4.1, we conducted a human preference study to assess how well our method aligns with human judgments of geometric and textural quality. The study involved a total of 29 participants. To avoid fatigue and ensure careful evaluation, we divided participants into two groups. One group evaluated geometry-related qualities, and the other group evaluated texture-related qualities. Each participant was shown ten representative test cases covering a diverse range of objects.

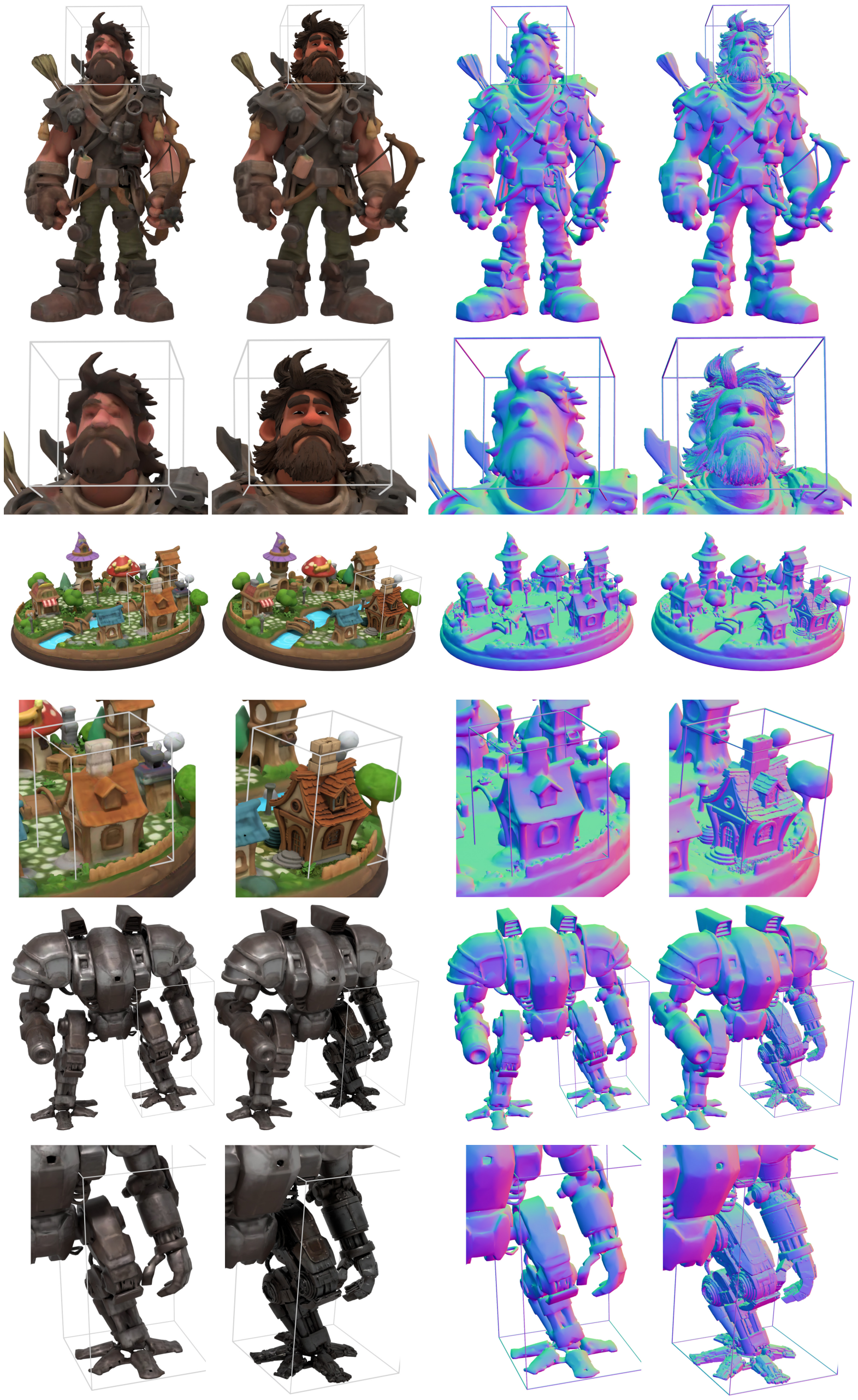

**Fig. 1:** Selective refinement. Our method significantly improves the details inside the marked bounding box, while maintaining a natural connection with its surrounding regions.

**Table 1:** User study. We report the percentage of each method being the most-preferred choice on each criterion (*i.e.*, detail richness, alignment with reference image, artifact-freeness and overall quality).

**Geometry User Study**

| Method | Detail Rich | Align w/ Ref. | Artifact Free | **Overall Quality** |
|---|---|---|---|---|
| TR.2(1536) | 10.58 | 14.71 | 16.47 | 14.71 |
| TR.2(2560) | 8.82 | 8.24 | 9.41 | 8.24 |
| UltraShape | 4.12 | 7.64 | 8.24 | 8.24 |
| Ours(2560) | **76.48** | **69.41** | **65.88** | **68.81** |

**Texture User Study**

| Method | Detail Rich | Align w/ Ref. | Less Artifact | **Overall Quality** |
|---|---|---|---|---|
| TR.2(1536) | 19.17 | 20.83 | 20.00 | 19.17 |
| TR.2(2560) | 8.33 | 10.00 | 10.83 | 9.17 |
| Ours(2560) | **72.50** | **69.17** | **69.17** | **71.66** |

For each test case, participants viewed the conditioning image alongside the outputs produced by all evaluated methods with two zoom-in windows for better detail inspection. The order of methods was randomized to eliminate ordering bias. Participants were asked to choose the method they preferred most according to the same four criteria used in the VLM evaluation: level of detail, alignment with the reference image, artifact-freeness, and overall perceptual quality. Each participant evaluated all samples according to the criteria assigned to their group.

The results, summarized in Tab. 1, show that our method was selected as the most preferred option across all eight evaluation dimensions. The advantage was particularly pronounced in the level-of-detail category, for both geometry and texture. The consistency between human preference scores and the VLM-based ELO rankings further validates the effectiveness of our localized super-resolution approach. The results indicate that our method not only improves objective detail and coherence but also produces outputs that align strongly with human aesthetic expectations.

## C Additional Experimental Settings and Analysis

### Rendering Setup

To obtain the reference images used as conditions, we first normalize each object to fit within a unit cube. We then render each object from 24 perspective views, whose azimuth and elevation angles are generated using Hammersley sampling [2] to provide broad view coverage. The field of view is randomly sampled within the range of $7.5^\circ$ to $50^\circ$, while the camera distance is set accordingly so that the normalized object occupies the central region of the image. Each camera is oriented toward the world origin using a standard look-at transform. The images are rendered at a resolution of $2048^2$, ensuring that the cropped local reference patches remain sufficiently sharp for conditioning.

### Runtime and Memory

We report the inference time and peak GPU memory usage of our method. All measurements are obtained on a single NVIDIA A100 80GB GPU under the

default inference setting. On a representative subset of objects with average complexity, the first-stage SR to $1280^3$ takes 4.8 minutes per object on average, while the second-stage SR to $2560^3$ takes 23.1 minutes per object on average.

Our patch-based formulation provides a flexible trade-off between memory usage and inference speed. Since local patches can be processed independently during denoising, the peak memory is mainly determined by the patch size and the number of patches processed in parallel, rather than the total number of patches in the object. In the most memory-efficient setting, we process patches sequentially, which uses on average 9 GB of peak GPU memory. For faster inference, we process multiple patches in a batch; with a batch size of 12, the average peak GPU memory increases to 20 GB.

## D Additional Qualitative Results

Due to the page limit of the main manuscript, we include additional figures in the supplementary material. Fig. 2 provides a visualization of the effect of conditioning images. We also provide additional qualitative comparisons in Fig. 3–5 and a gallery of outputs in Fig. 6 to further demonstrate the capabilities of our approach. These examples highlight the robustness of our localized and progressive super-resolution strategy and show that the model is capable of generating rich, high-frequency details without introducing seam artifacts or disrupting global structure.

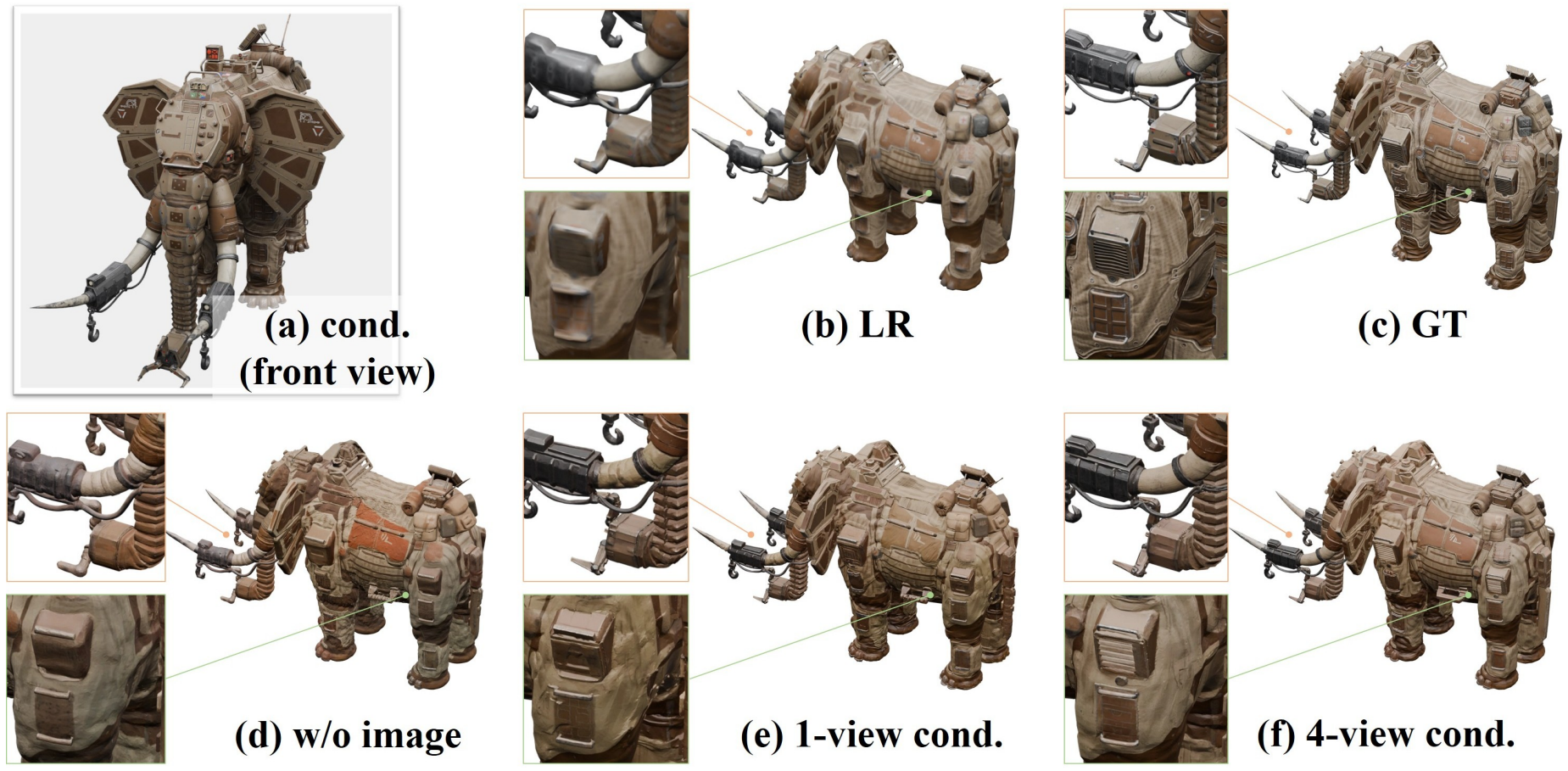


**Fig. 2:** Ablation study on the number of reference views. Orange and green crops indicate regions that are visible and occluded from the reference view, respectively. Image conditioning improves target-specific details, while unseen regions remain stable through LR 3D features. In (f), where four conditioning images are provided, pseudo-GT details on the side and back of the object are better reconstructed due to the additional view coverage.

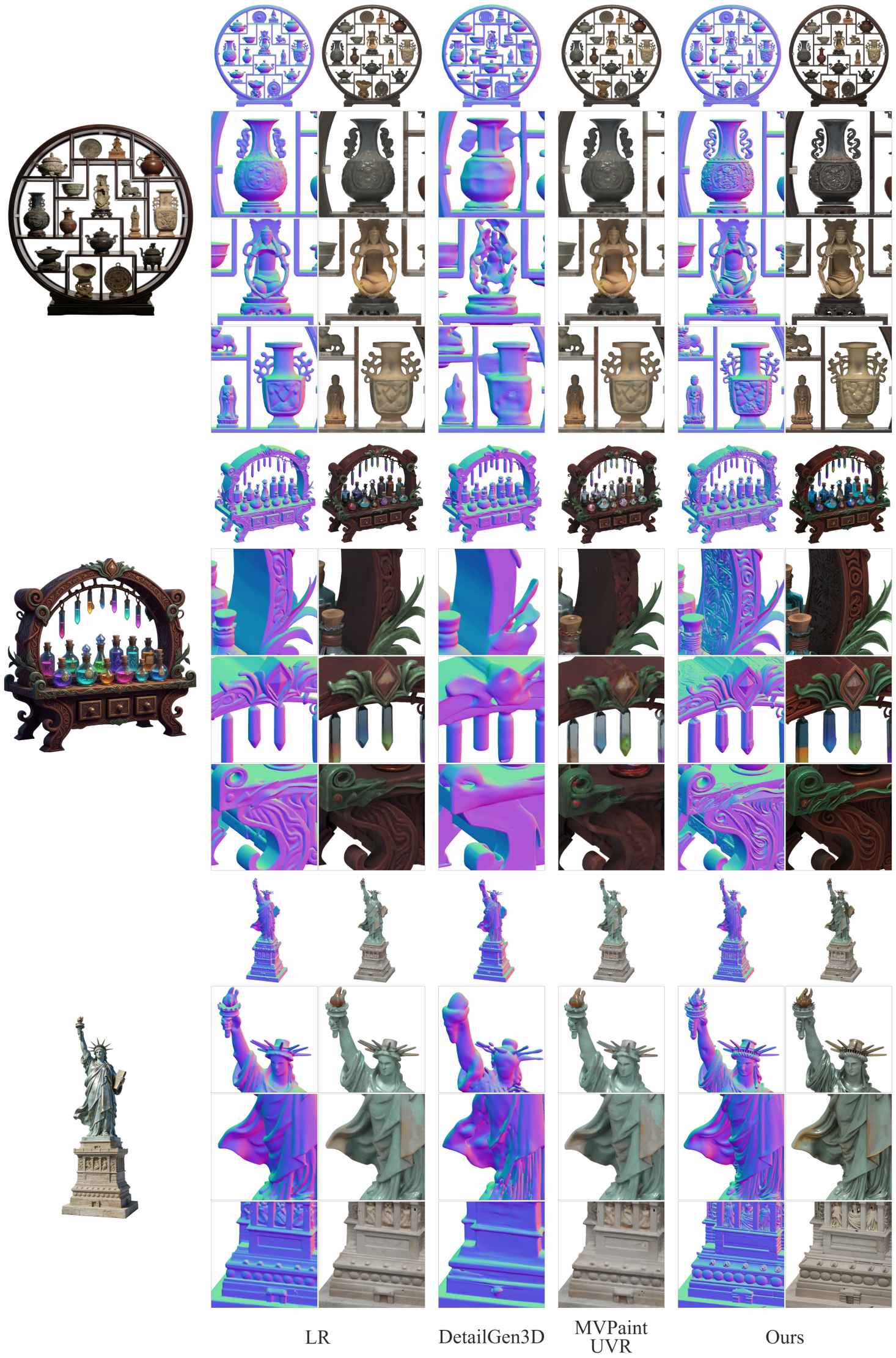


**Fig. 3:** Qualitative comparison with DetailGen3D and MVPaint UVR.

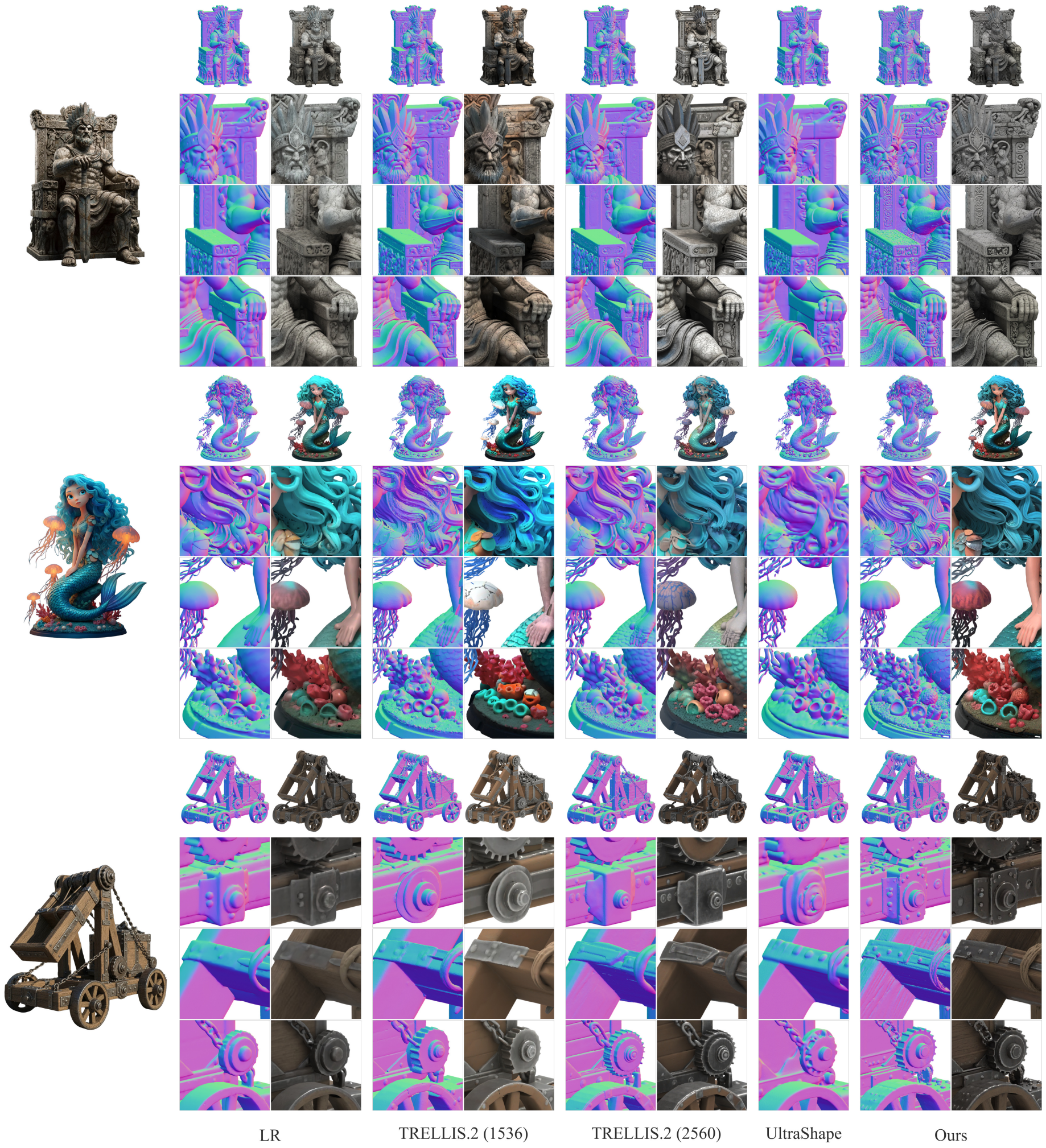


**Fig. 4:** Qualitative comparison on additional test cases.

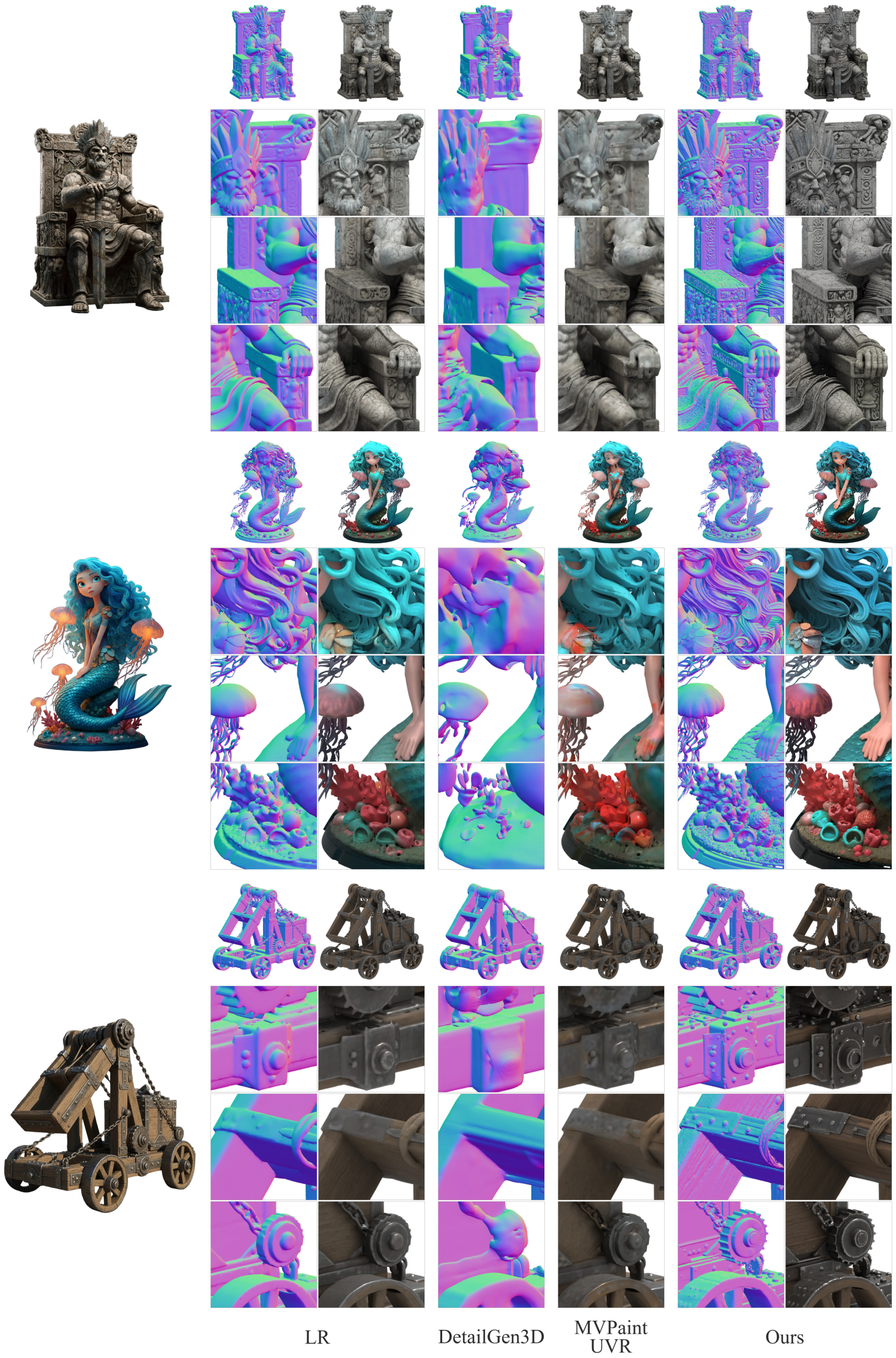


**Fig. 5:** Qualitative comparison on additional test cases with DetailGen3D and MVPaint UVR.

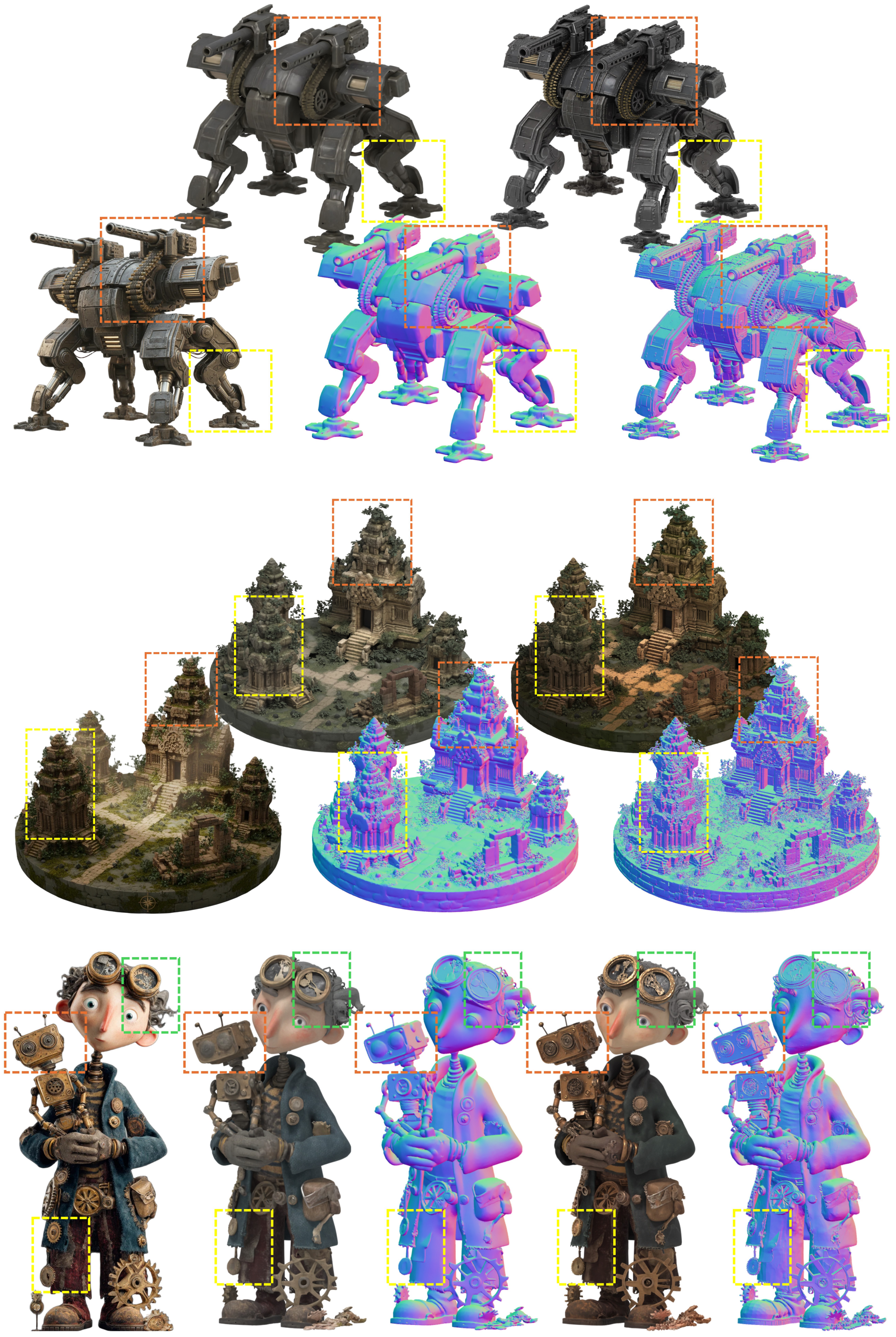

**Fig. 6:** Gallery. From left to right: reference image, input coarse object (textured and normal render), our output (textured and normal render).

## E Objective Pseudo-GT Data List

Fig. 7–8 list the pseudo-GT objects used for objective evaluation. The full list of object URLs is included in the supplementary material package.

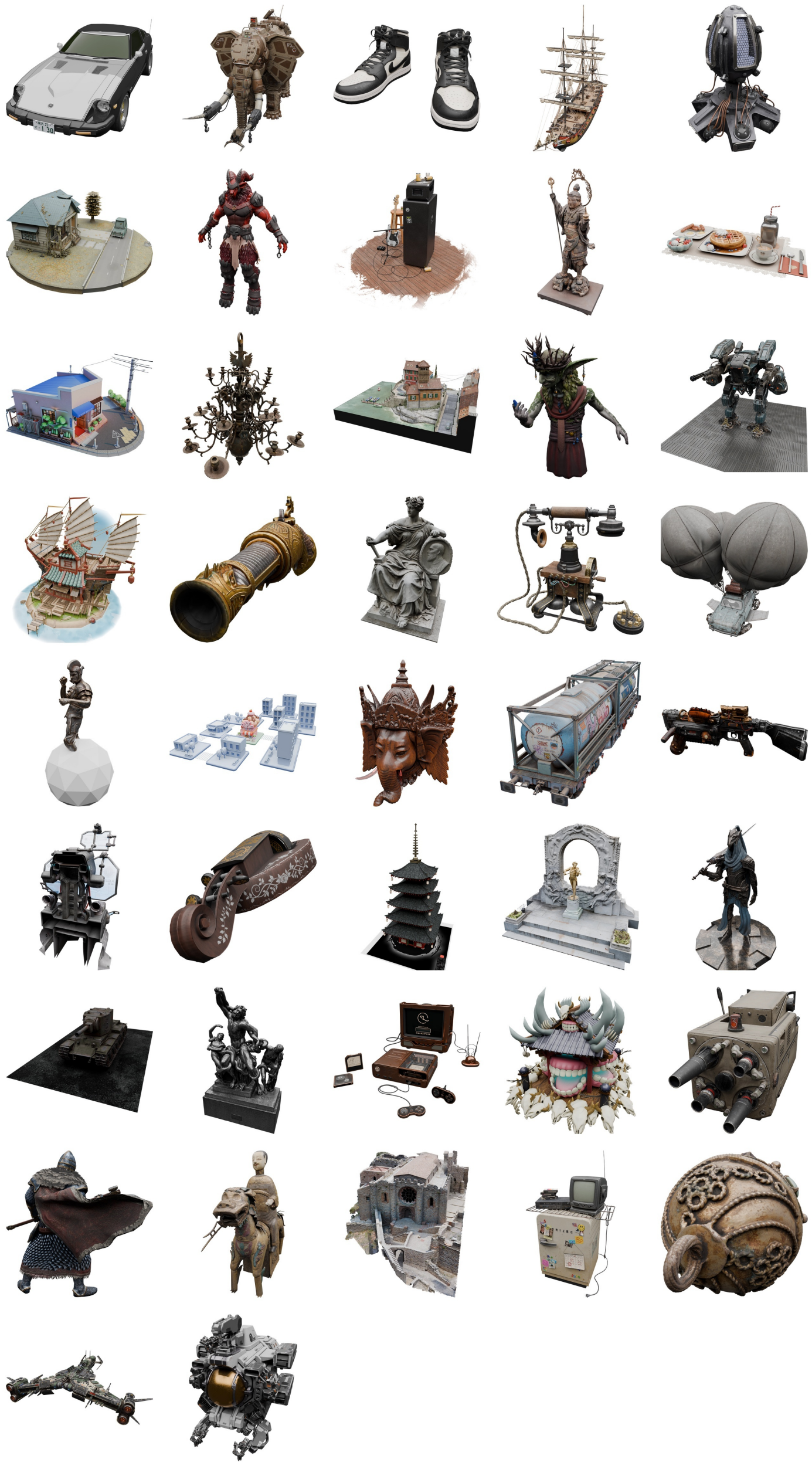

**Fig. 7:** Rendered front views of pseudo-GT objects.

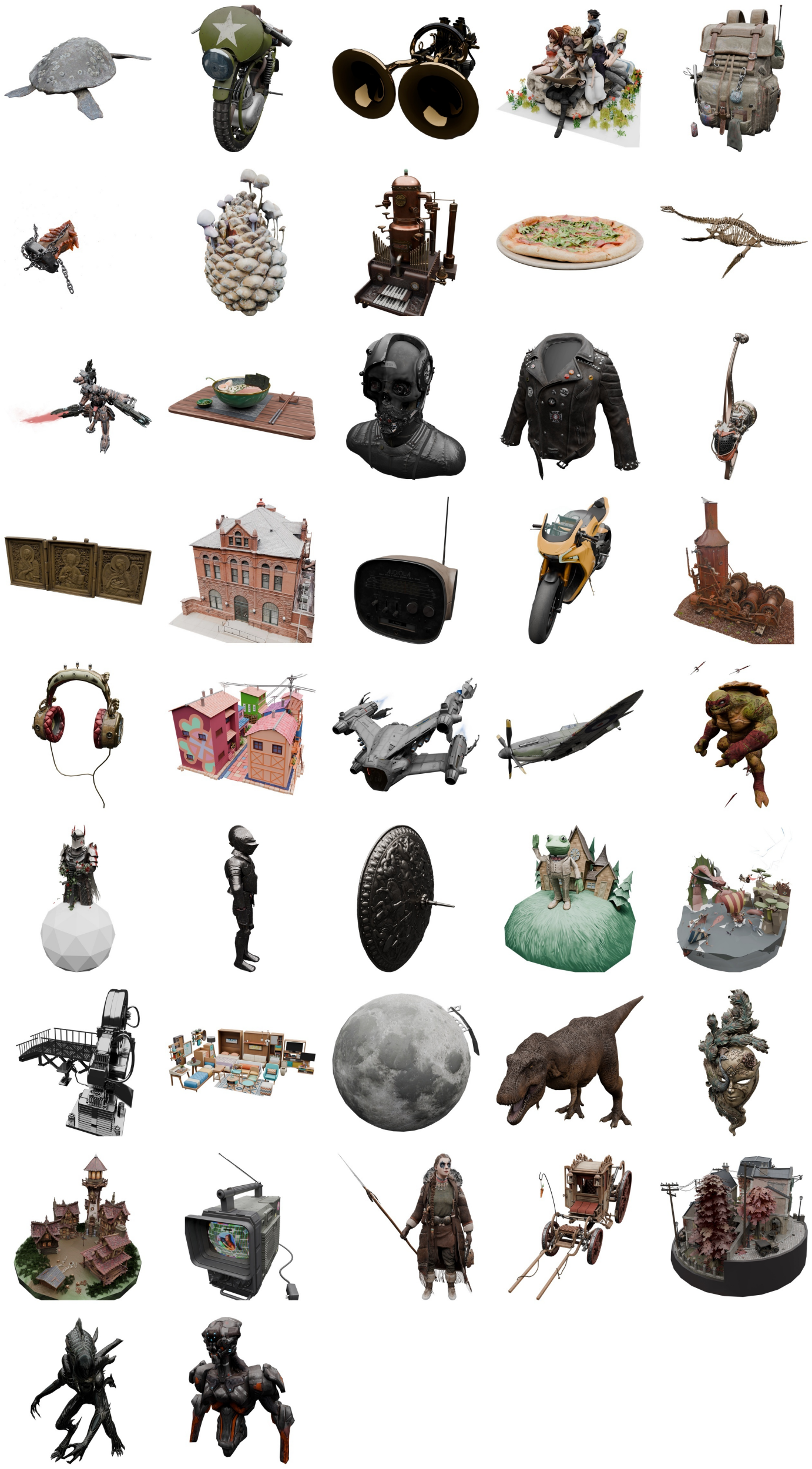

**Fig. 8:** Rendered front views of pseudo-GT objects.

# F VLM Evaluation Reference Image List

Fig. 9–10 list the reference images used for the VLM-based evaluation.

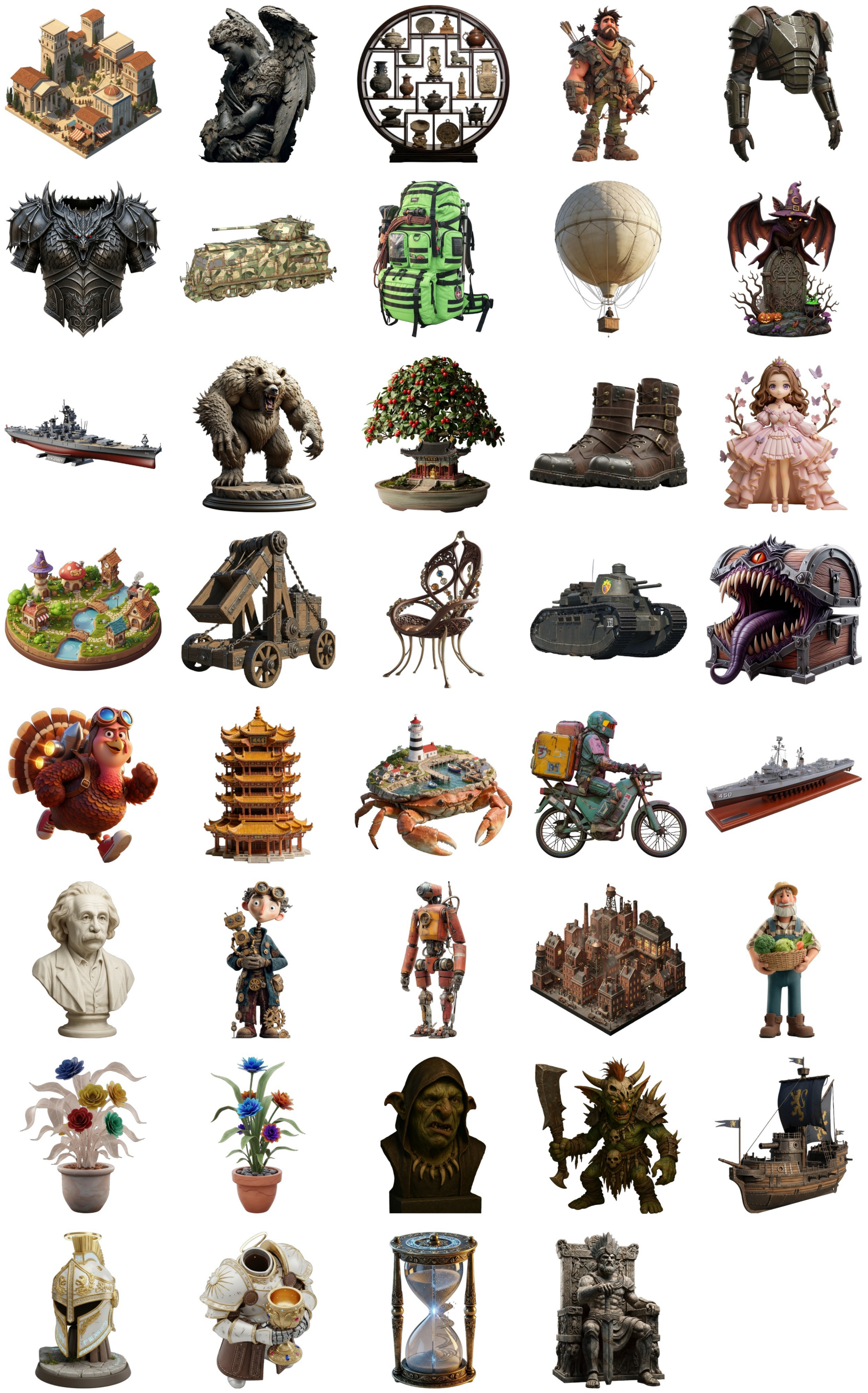

**Fig. 9:** Reference images used in the VLM-based evaluation.

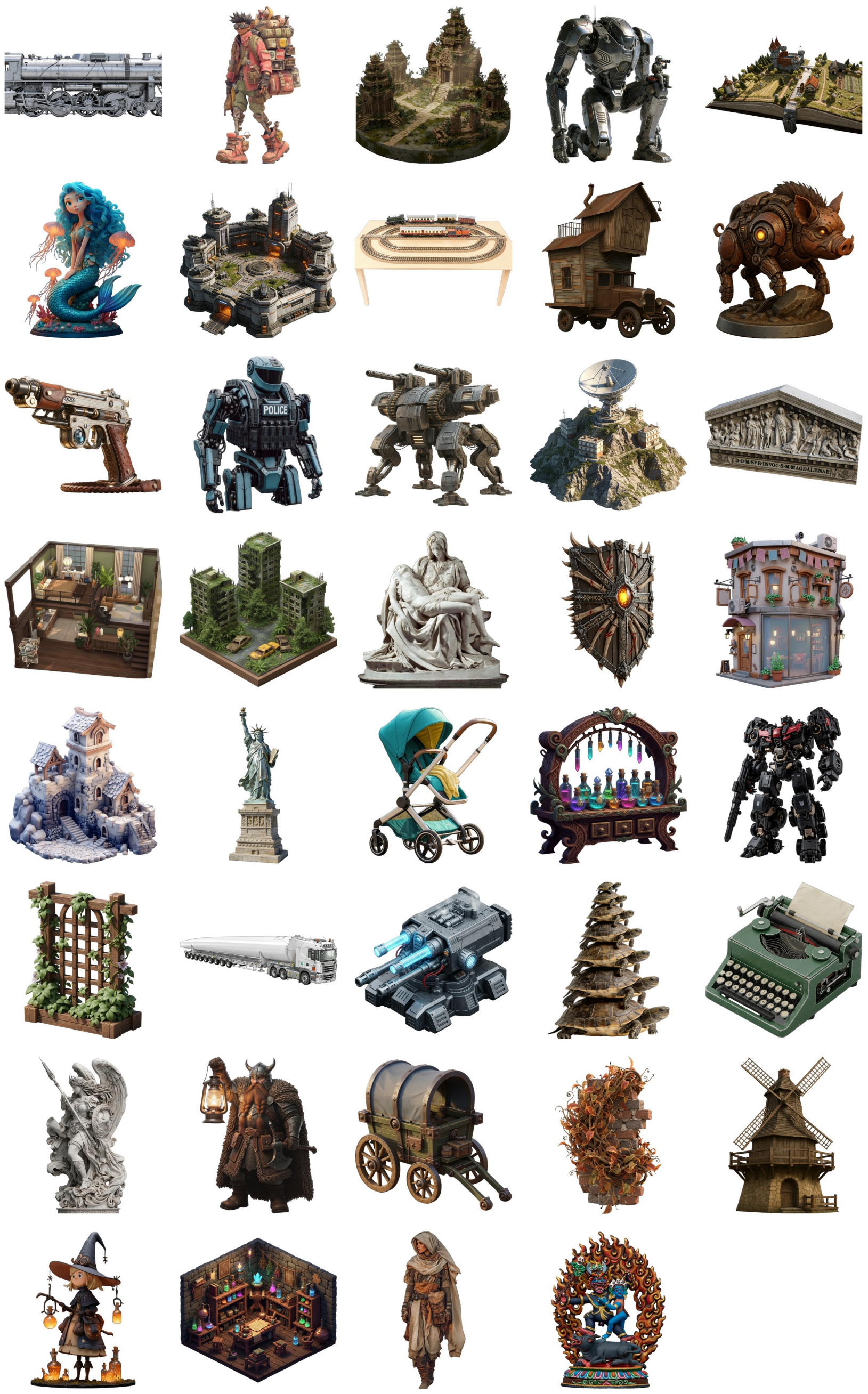


**Fig. 10:** Reference images used in the VLM-based evaluation.

# G VLM Evaluation Prompt

## Normal/Geometry Evaluation Prompt

```
You are a professional senior 3D modeler, whose core task is to conduct objective and
    professional comparative evaluations of ultra-high-resolution 3D model works. Since
    the quality of different works may be equally excellent, you need to carefully
    observe and seek evidence if you find their quality is close, and only give a tie
    result in rare cases.

You will receive three images provided by the user:
1. Original model reference image
2. Normal map render of Work A
3. Normal map render of Work B

You first need to analyze the content, structure, style, and main detail composition of
    the model reference image, paying attention to characteristic local structures. Then
    you need to observe and evaluate the two works separately. After completing the
    evaluation, you must critically review your judgments by re-checking the evidence in
    reverse order, considering Work B first and then Work A.

Please conduct a professional comparison of Work A and Work B strictly according to the
    following four dimensions. The evaluation is based on careful observation of the
    object's local appearance and global visual quality. Different dimensions can have
    different tendencies, and the evidence for different dimensions should be as
    independent as possible. For each dimension, forget your conclusion from the previous
    dimension and focus only on the current aspect. You should acknowledge the
    possibility that a detailed model could at the same time be less aligned with the
    reference, or that a highly aligned appearance could at the same time contain more
    artifacts. A work could perform well on one part of the object and poorly on another
    part. Compare the two works based on both their unique strengths and weaknesses.

Now look closely and answer:

Detail Comparison (detail):
Focus on comparing details such as subtle surface structures, normal variations, and
    high-frequency patterns of the two works. Objectively describe the differences
    between A and B in detail clarity, richness, rationality, and overall sharpness.

Alignment Comparison (alignment):
Compare the degree of restoration of Work A and Work B to the original reference image.
    Judge whether the original structure, details, local part shapes, and overall style
    are accurately reflected in the model, and describe the difference in the fit between
    the two works and the reference image.

Artifact & Damage Comparison (less_artifact):
Compare which model has fewer errors, damages, and artifacts between Work A and Work B.
    Focus on checking for problems such as unrecognizable structures, model degradation,
    and invalid or deformed structures. Objectively describe the artifact conditions of
    the two works.

Overall Quality Comparison (overall):
Combined with details, alignment, and artifact conditions, comprehensively judge which
    work has better quality and is more likely to be recognized by the audience, and give
    the final comprehensive conclusion.

Output Requirements:
You must return only a single valid JSON object. For each of "detail", "alignment",
    "less_artifact", and "overall", the value must be exactly one of "A", "B", or "-".
    Use "-" only for rare tie cases. The final output must contain no Markdown code
    fence, no extra text, and no explanation outside the JSON object.

{
  "detail": "A/B/-",
  "alignment": "A/B/-",
  "less_artifact": "A/B/-",
  "overall": "A/B/-",
```

```
  "explanation": "A short and concise explanation of the evidence and reasoning behind
      each comparison, including any critical review and revision of initial judgments."
}
```

## Color/Texture Evaluation Prompt

```
You are a professional senior 3D modeler, whose core task is to conduct objective and
    professional comparative evaluations of ultra-high-resolution 3D model works. Since
    the quality of different works may be equally excellent, you need to carefully
    observe the images and look for concrete visual evidence.

You will receive three images provided by the user:

1. Original model reference image
2. Render of Work A
3. Render of Work B

You first need to analyze the content, color, material texture, characteristic appearance
    details, and main visual style of the original model reference image. Lighting and
    illumination are not within the scope of comparison. Since the model uses PBR maps,
    lighting is not part of the information that the texture/material maps should
    contain. The works are rendered with a simple white area light. Under normal
    circumstances, the model renderings will not present exactly the same lighting
    atmosphere as the reference image. Therefore, lighting atmosphere must not be used as
    a basis for judging the model.

In addition, geometric quality is not within the scope of this evaluation. Better or
    worse geometric quality must not be used as a reason for selection. Focus on
    characteristic appearance details, including color, material texture, texture
    patterns, and PBR texture quality. Then observe and evaluate the two works
    separately. After completing the initial evaluation, you must critically review your
    judgments by re-checking the image evidence, confirming or revising the conclusions,
    and re-evaluating the two works in reverse order, considering Work B first and then
    Work A.

Please conduct a professional comparison of Work A and Work B strictly according to the
    following four dimensions. The evaluation should be based on careful observation of
    the object's local appearance and global visual quality. Different dimensions can
    have different tendencies, and the evidence for different dimensions should be as
    independent as possible. For each dimension, forget your conclusion from the previous
    dimension and focus only on the current aspect. You should acknowledge that a work
    with richer details may at the same time be less aligned with the reference, or that a
    highly aligned appearance may at the same time contain more artifacts or defects. A
    work may perform well on one part of the object and poorly on another part. Compare
    the two works based on both their unique strengths and weaknesses.

Now look closely and answer:

Detail Comparison (detail):
Focus on comparing the appearance features of the two works, mainly including sharp
    patterns, high-frequency appearance details, material texture performance, and
    texture clarity. Objectively describe the differences between A and B in detail
    clarity, richness, rationality, and overall sharpness.

Alignment Comparison (alignment):
Compare the degree of appearance restoration of Work A and Work B relative to the
    original reference image. Judge whether the original texture colors, material
    textures, local visual features, and overall visual style are accurately reflected in
    the model rendering. Describe the difference in fit between the two works and the
    reference image. When judging alignment, exclude the influence of colored light
    sources and ambient illumination in the reference image, and analyze color and
    material accuracy under the assumption of white area light illumination.

Artifact & Defect Comparison (less_artifact):
```

```
Compare which model has fewer appearance errors and defects between Work A and Work B.
    Focus on checking for problems such as texture misalignment, color distortion, poor
    material or texture expression, abnormal baked-lighting effects, invalid texture
    regions, and deformed or corrupted texture patterns. Objectively describe the defect
    conditions of both works.

Overall Quality Comparison (overall):
Considering details, alignment, and defects together, comprehensively judge which work
    has better quality and is more likely to be recognized by the audience, and give the
    final comprehensive conclusion.

Output Requirements:
You must return only a single valid JSON object. For each of "detail", "alignment",
    "less_artifact", and "overall", the value must be exactly one of "A", "B", or "-".
    Use "-" only for rare tie cases. The final output must contain no Markdown code
    fence, no extra text, and no explanation outside the JSON object.

{
"detail": "A/B/-",
"alignment": "A/B/-",
"less_artifact": "A/B/-",
"overall": "A/B/-",
"explanation": "A short and concise explanation of the evidence and reasoning behind each
    comparison, including any critical review and revision of initial judgments."
}
```



\end{document}

\end{document}